\documentclass[11pt]{article}

\usepackage[final]{acl}

\usepackage{times}
\usepackage{latexsym}
\usepackage{amsmath}
\usepackage{mathtools}
\usepackage{amsthm}
\usepackage{breqn}
\usepackage{caption}
\usepackage[T1]{fontenc}

\usepackage[utf8]{inputenc}

\usepackage{microtype}

\usepackage{inconsolata}

\usepackage{graphicx}

\usepackage{algorithm}
\usepackage{algpseudocode}
\usepackage{listings}
\usepackage{hyperref}
\usepackage{geometry}
\usepackage{xcolor}
\usepackage[table]{xcolor}
\usepackage{multirow}
\usepackage{amssymb}
\usepackage[normalem]{ulem}
\usepackage{tcolorbox}
\usepackage{subcaption}
\tcbuselibrary{breakable}
\usepackage{booktabs}
\usepackage{tabularx}
\usepackage{rotating} 
\tcbuselibrary{listings, skins, breakable}
\newtcbox{\codebox}{colback=gray!5, colframe=gray!30, boxrule=0.5pt, arc=2pt, left=6pt, right=6pt, top=6pt, bottom=6pt, breakable}

\hypersetup{
    colorlinks=true,
    linkcolor=blue,
    filecolor=magenta,      
    urlcolor=magenta,
}

\definecolor{promptbgcolor}{RGB}{245,245,245} 
\definecolor{promptframecolor}{RGB}{80,80,80} 

\newtcblisting{promptbox}[1][]{
  enhanced,
  colback=promptbgcolor,     
  colframe=promptframecolor, 
  boxrule=0.5pt,             
  arc=2mm,                   
  width=\linewidth,          
  listing only,              
  listing options={
    basicstyle=\scriptsize\ttfamily, 
    breaklines=true,                 
    breakatwhitespace=false,         
    breakindent=0pt,                 
    columns=flexible,                
    keepspaces=true,                 
    showstringspaces=false,          
    escapechar=|                     
  },
  title={\bfseries\sffamily\small #1}, 
  #1
}

\definecolor{codebgcolor}{RGB}{245,245,245}
\definecolor{codeframecolor}{RGB}{80,80,80}
\definecolor{codegreen}{rgb}{0,0.6,0}
\definecolor{codegray}{rgb}{0.5,0.5,0.5}
\definecolor{codepurple}{rgb}{0.58,0,0.82}

\newtcblisting{mycodebox}[1][]{
  enhanced,                  
  colback=codebgcolor,       
  colframe=codeframecolor,   
  boxrule=0.5pt,             
  arc=2mm,                   
  width=0.98\textwidth,      
  center,                    
  listing only,              
  listing options={
    language=Python,                 
    basicstyle=\ttfamily\footnotesize, 
    breaklines=true,                 
    columns=fullflexible,            
    keepspaces=true,                 
    showstringspaces=false,          
    tabsize=4,                       
    commentstyle=\color{codegreen},
    keywordstyle=\color{magenta},
    stringstyle=\color{codepurple},
    gobble=0                         
  }
}
\title{Beyond Static Tools: Test-Time Tool Evolution for Scientific Reasoning}

\author{
    \bfseries
    Jiaxuan Lu$^{1,\dagger}$, 
    Ziyu Kong$^{2,\dagger}$, 
    Yemin Wang$^{3,\dagger}$, 
    Rong Fu$^{4}$, 
    Haiyuan Wan$^{1,5}$, \\
    \bfseries
    Cheng Yang$^{6}$, 
    Wenjie Lou$^{1}$, 
    Haoran Sun$^{1}$, 
    Lilong Wang$^{1}$, \\
    \bfseries
    Yankai Jiang$^{1}$, 
    Xiaosong Wang$^{1}$, 
    Xiao Sun$^{1}$, 
    Dongzhan Zhou$^{1,*}$ \\
    \\
    $^1$Shanghai Artificial Intelligence Laboratory \quad
    $^2$Fudan University \\
    $^3$Xiamen University \quad
    $^4$University of Macau \quad
    $^5$Tsinghua University \\
    $^6$Hangzhou Dianzi University \\
    \\
    
    \footnotetext[2]{Equal contribution.} 
    \footnotetext[1]{Corresponding author.} 
}

\begin{document}
\maketitle

\begin{figure*}[t]
  \centering
  \includegraphics[width=\textwidth]{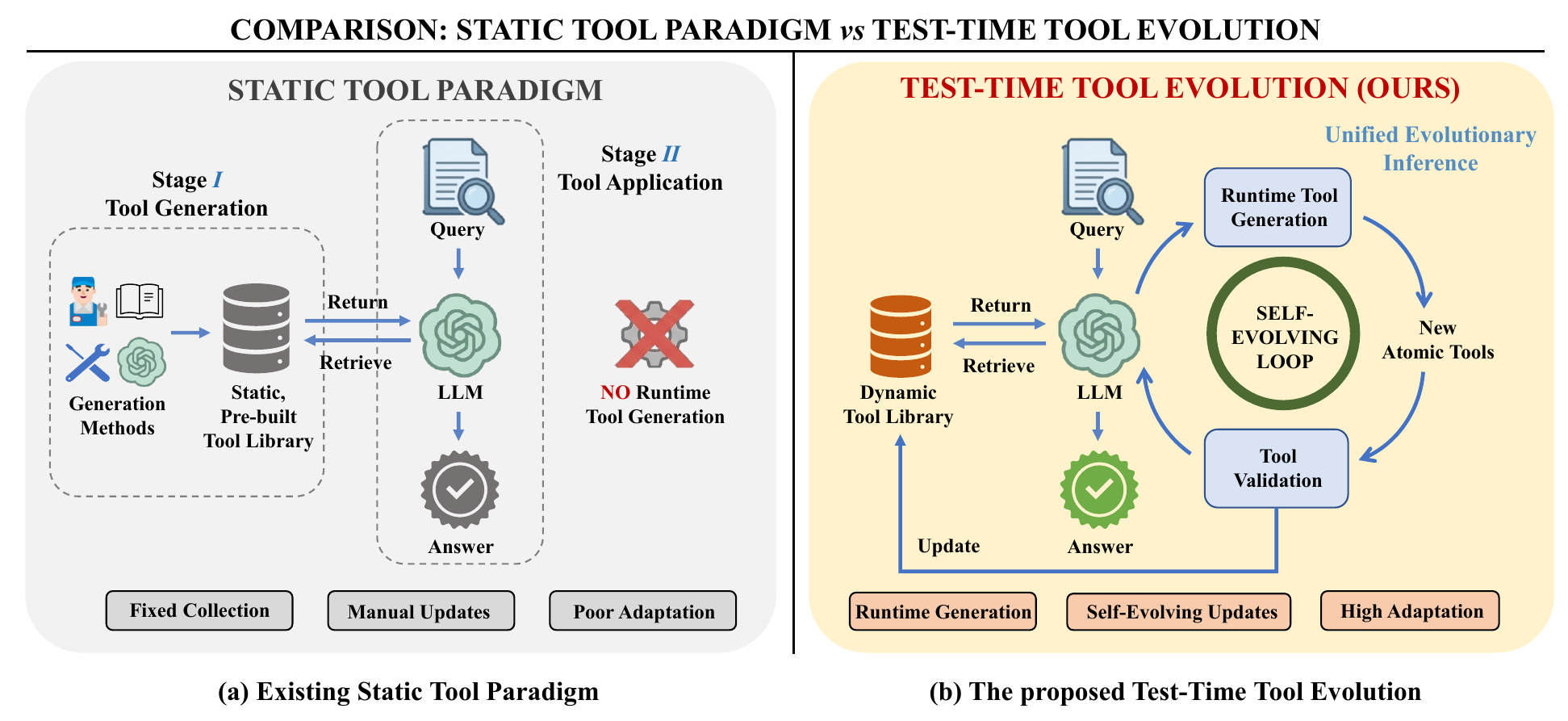}
  \caption{Paradigm comparison: Static Tool Paradigm (left) vs Test-Time Tool Evolution (right). Static approaches require pre-collected tool libraries, limiting coverage and domain adaptability. Our test-time evolution starts with an empty library and generates tools on-demand during problem-solving, enabling continuous evolution to new domains and problems.}
  \label{fig:teaser}
\end{figure*}

\begin{abstract}
The central challenge of AI for Science is not reasoning alone, but the ability to create computational methods in an open-ended scientific world. Existing LLM-based agents rely on static, pre-defined tool libraries, a paradigm that fundamentally fails in scientific domains where tools are sparse, heterogeneous, and intrinsically incomplete. In this paper, we propose Test-Time Tool Evolution (TTE), a new paradigm that enables agents to synthesize, verify, and evolve executable tools during inference. By transforming tools from fixed resources into problem-driven artifacts, TTE overcomes the rigidity and long-tail limitations of static tool libraries. To facilitate rigorous evaluation, we introduce SciEvo, a benchmark comprising 1,590 scientific reasoning tasks supported by 925 automatically evolved tools. Extensive experiments show that TTE achieves state-of-the-art performance in both accuracy and tool efficiency, while enabling effective cross-domain adaptation of computational tools. The code and benchmark have been released at \url{https://github.com/lujiaxuan0520/Test-Time-Tool-Evol}.
\end{abstract}

\section{Introduction}
The ultimate pursuit of ``AI for Science'' is to construct autonomous agents capable of navigating the unbounded complexity of the physical world, from discovering novel drug candidates to deriving governing equations of matter. While Large Language Models (LLMs) act as powerful reasoning engines \citep{brown2020language},  
scientific research demands precise, executable rigor that inherently exceeds the probabilistic nature of LLMs \citep{miret2024llms}. Without the mediation of computational tools, scientific LLMs demonstrate significantly limited performance \citep{yu2025tooling}, hallucinating on tasks requiring rigorous fidelity \citep{chen2021evaluating,nijkamp2022conversational}.

Current paradigms attempt to bridge this gap through static tool libraries, \textit{i.e.}, pre-defined functions constructed via manual curation or offline synthesis. While effective for standardized tasks (\textit{e.g.}, weather, booking), this paradigm collapses in scientific reasoning. 
We identify two fatal bottlenecks in the static approach. First, scientific tools exhibit extreme sparsity and heterogeneity. Unlike the abundant ecosystems of general domains, scientific functions are scattered and non-standardized, rendering the manual curation of a comprehensive library computationally intractable.
Second, and most critically, static libraries cannot anticipate the bespoke computational primitives required for novel inquiry, which confines agents to the role of passive selectors rather than active discoverers, imposing an artificial ceiling on their potential to solve unseen problems \citep{schick2023toolformer,wan2025deepresearch}.

We contend that scientific reasoning is fundamentally unsuited for the static tool paradigm shown in Figure~\ref{fig:teaser}(a). For an agent to be a genuine scientist, it cannot merely \emph{select} tools, it must \emph{evolve} them. Unlike existing approaches using limited pre-defined tool libraries, we propose Test-Time Tool Evolution (TTE), a paradigm shift that transitions scientific reasoning from static retrieval to dynamic evolution.
Instead of relying on a fossilized library, TTE synthesizes executable tools on demand during the inference phase. By dynamically decomposing complex problems into atomic functions and verifying them in real-time, TTE ensures that every tool in the library is intrinsically aligned with the problem space.
We instantiate TTE in two fundamental tasks: Ab-initio Tool Synthesis (TTE-Zero) where the agent evolves a tool library from scratch to solve problems without prior knowledge, and Cross-Domain Tool Adaptation (TTE-Adapt) where the agent dynamically repurposes a source tool library (\textit{e.g.}, Materials Science) to conquer a new domain (\textit{e.g.}, Chemistry).

Our contributions are summarized as follows:
\begin{enumerate}
\item We introduce \textbf{Test-Time Tool Evolution}, a novel framework that mirrors the iterative nature of the scientific method.
By enabling tools to be generated, verified, and evolved during inference, TTE overcomes the inherent rigidity of static paradigms.
\item We release \textbf{SciEvo}, a comprehensive benchmark for evaluating tool evolution, comprising 1,590 scientific evaluation instances supported by a library of 925 evolved tools. 
\item Extensive evaluations demonstrate that TTE establishes a new State-of-the-Art (SOTA) for scientific reasoning. Specifically, TTE-Zero outperforms existing baselines in both accuracy and tool utilization efficiency significantly, while TTE-Adapt enables effective cross-domain tool adaptation, demonstrating the transferability of computational primitives across scientific disciplines.

\end{enumerate}

\section{Related Work}

\subsection{Static Tool Paradigm}
The paradigm of augmenting LLMs with external tools has expanded their capabilities beyond static parametric knowledge. Foundational works have established the mechanisms for this interaction, \textit{e.g.}, ReAct \citep{yao2022react} introduces the interleaving of reasoning traces with tool actions, while Toolformer \citep{schick2023toolformer} demonstrates that LLMs could teach themselves to use calculator and search APIs via self-supervised fine-tuning. Building on these execution frameworks, subsequent research have focused on scaling the tool space. Systems like Gorilla \citep{patil2024gorilla} and ToolLLM \citep{qin2023toolllm} employ instruction tuning and retrieval-based mechanisms to select appropriate tools from massive, pre-defined API libraries, \textit{e.g.}, HuggingFace or RapidAPI, enabling models to address diverse general-domain queries.

The static tool paradigm has been widely adapted to specialized scientific domains to address the complexity of domain-specific tasks. In chemistry and materials science, systems like ChemCrow \citep{bran2023chemcrow}, CheMatAgent \citep{wu2025chemagent}, and ChemMAS \citep{yang2025multi} integrate fixed sets of expert-curated tools ranging from simple calculators to complex synthesis planners to automate organic synthesis and drug discovery. Other approaches focus on enhancing domain capability through knowledge-base integration, such as HoneyComb \citep{zhang2024honeycomb}, or utilizing multi-agent frameworks to uncover hidden interdisciplinary relationships as explored in SCP \citep{jiang2025scp}.
Finally, recent works rigorously benchmark the impact of these static toolsets, as seen in ChemToolAgent \citep{yu2025tooling} and MatTools \citep{liu2025mattools}.
Despite their effectiveness in bounded scenarios, these systems share a critical limitation, \textit{i.e.}, they rely on pre-defined, static tool libraries, which fail to exhaustively cover the open-ended task space.

\subsection{Dynamic Tool Synthesis}

To address the coverage limitations of static libraries, recent research has shifted towards enabling LLMs to generate tools dynamically. Approaches such as CREATOR \citep{qian2023creator} and CRAFT \citep{yuan2024craft} leverage the code generation capabilities of LLMs to synthesize custom tools via abstract reasoning to solve specific problems. However, these methods typically treat tool generation as a one-off process or, as seen in LATM \citep{cai2023large} and ToolMaker \citep{wolflein2025llm}, adopt a decoupled paradigm where the tool-making phase is separated from inference, hindering real-time adaptation.

Moving beyond static generation, systems like Voyager \citep{wang2023voyager} introduce the concept of an evolving skill library, allowing agents to accumulate executable code as tools through trial and error in embodied environments. Similarly, SEAgent \citep{sun2025seagent} and ToolACE-DEV \citep{huang2025toolace} investigate self-evolving mechanisms for operating system control. While promising, these evolutionary frameworks are designed for gamified or general computer tasks, lacking the rigor and domain-specific logic required for scientific reasoning. Parallelly, automated design approaches \citep{hu2024automated,shang2024agentsquare} explore searching for optimal agent architectures within modular design spaces, focusing on the arrangement of components rather than the evolution of the tools themselves.

\begin{figure*}[t]
  \centering
  \includegraphics[width=\textwidth]{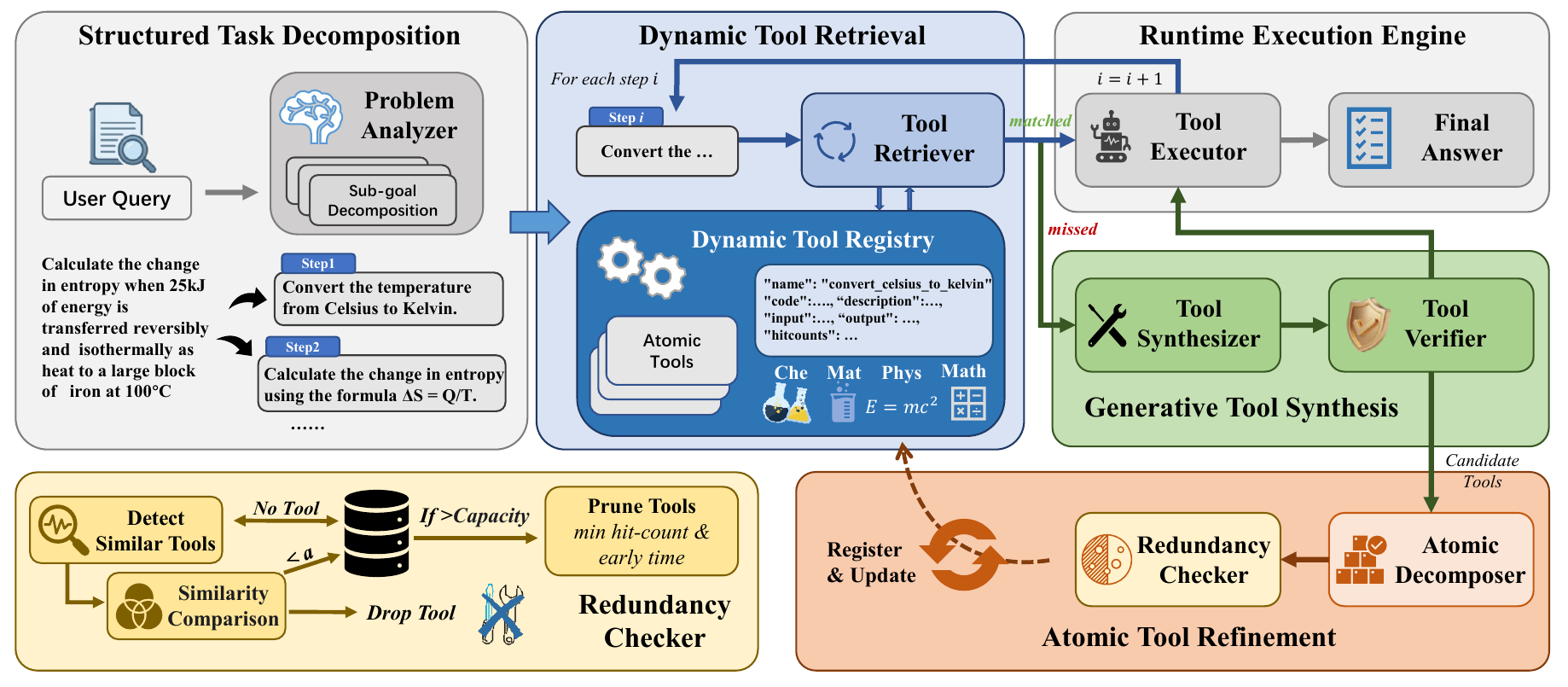}
    \caption{The architecture of the Test-Time Tool Evolution (TTE) framework. The system operates through a closed-loop workflow comprising five integrated stages. (1) Structured Task Decomposition: The Problem Analyzer decomposes complex scientific queries into a sequence of executable sub-goals. (2) Dynamic Tool Retrieval: The system queries the Dynamic Tool Registry for existing atomic tools. If retrieval fails, it triggers (3) Generative Tool Synthesis: The Tool Synthesizer creates candidate tools on-the-fly, which undergo strict verification by the Tool Verifier. (4) Atomic Tool Refinement: Validated tools are decoupled into reusable atomic units by the Atomic Decomposer, filtered by the Redundancy Checker, and registered to update the library. (5) Runtime Execution Engine: Once the required tools are successfully retrieved or generated for all the steps, the Tool Executor executes the sequence to synthesize the final answer.}
  \label{fig:architecture}
\end{figure*}

\section{Test-Time Tool Evolution}
\subsection{Problem Definition}
We formalize \emph{Test-Time Tool Evolution} (TTE) as a fundamentally new paradigm that addresses a critical gap in existing static tool paradigms. Unlike existing approaches where tools are prepared offline before problem-solving, TTE enables tools to be generated and evolved \emph{during} problem-solving, representing a paradigm shift from static to dynamic tool ecosystems.

Formally, given a sequence of scientific problems $\mathcal{P} = \{P_1, P_2, \ldots, P_t\}$ arriving sequentially at test time, the goal of TTE is to maintain an evolving tool library $L_t$ that balances capability and efficiency. We frame this task as an online optimization problem where the system seeks to maximize the cumulative utility:
\begin{equation}
\label{eq:objective}
\max_{\{L_t\}_{t=1}^T} \sum_{t=1}^{T} \left( \mathbb{I}(\text{Solved}(P_t, L_t)) - \lambda \cdot |L_t| \right),
\end{equation}
where $\mathbb{I}(\cdot)$ is the indicator function for problem resolution, \textit{e.g.}, accuracy, and $\lambda$ is a regularization coefficient penalizing library expansion.
parameter training.
We instantiate TTE for two primary tasks: TTE-Zero for ab-initio tool synthesis ($L_0 = \emptyset$), and TTE-Adapt for cross-domain adaptation of a pre-defined tool library to a new target domain.
Since finding the global optimum for Eq.~\ref{eq:objective} is computationally intractable due to the combinatorial nature of tool composition, our TTE framework adopts a greedy evolution strategy. At each step $t$, the system updates $L_t$ to $L_{t+1}$ via tool generation and pruning mechanisms to approximate the optimal trajectory without explicit parameter training.

\subsection{Architecture Overview}
Our framework implements a closed-loop evolutionary workflow comprising five integrated modules, as shown in Figure~\ref{fig:architecture}. Structured Task Decomposition decomposes complex queries into executable sub-goals. Dynamic Tool Retrieval queries the library for existing tools. Generative Tool Synthesis creates new tools on-demand when retrieval fails. Atomic Tool Refinement decouples, validates, and registers new tools to evolve the library. Runtime Execution Engine executes the tool sequence to derive the final answer. The proposed architecture enables continuous library growth from an empty state while solving real-world problems.

\subsection{Structured Task Decomposition}
The \textit{Problem Analyzer} serves as the planning engine, decomposing scientific problems into a sequence of executable sub-goal operations. Given a problem $P$, it identifies the set of required operations $\mathcal{O}$:
\begin{equation}
\label{eq:problem-analysis}
\begin{aligned}
\mathcal{O} &= \text{Analyze}(P) \\
&= \{\, O_i : O_i \text{ is required to solve } P \,\}.
\end{aligned}
\end{equation}
The decomposition is tool-aware, isolating specific sub-goals that require computational intervention, setting the stage for the retrieval process.

\subsection{Dynamic Tool Retrieval}
For each identified operation $O_i$, the system queries the \emph{Dynamic Tool Registry}. We verify the existence of suitable tools using semantic similarity between their textual descriptions:
\begin{equation}
\label{eq:retrieval}
\text{sim}(O_i, T_j) = \cos(\text{embed}(d_{O_i}), \text{embed}(d_{T_j})),
\end{equation}
where $d$ denotes the functional description. The system makes a branching decision based on the maximum similarity score found in the current library $L$:
\begin{equation}
\label{eq:retrieval-decision}
T^* =
\begin{cases}
\displaystyle
\arg\max_{T_j \in L} \text{sim}(O_i, T_j), & \text{if } s_{\max} \ge \tau, \\[8pt]
\text{Generate}(O_i, P), & \text{otherwise,}
\end{cases}
\end{equation}
where $s_{\max} = \max_{T_j \in L} \text{sim}(O_i, T_j)$, $\tau$ is the threshold maximizing F1 score. The \textit{Tool Retriever} balances exploitation and exploration, ensuring efficient reuse of existing tools (the "matched" path) while automatically triggering the synthesis pipeline (the "missed" path) for novel requirements.

\subsection{Generative Tool Synthesis}
When retrieval fails, the Generative Tool Synthesis module creates a new tool through a rigorous generation-verification process.
Given $P$ and $O_i$, the \textit{Tool Synthesizer} proposes a tool $T_{\text{proposed}}$ via chain-of-thought reasoning:
\begin{equation}
\label{eq:generation}
P(T_{\text{proposed}} \mid P, O_i) = \prod_{k=1}^{K} P(f_k \mid P, O_i, f_{1:k-1}),
\end{equation}
where $f_k$ represents components such as function signature and implementation, and $K$ denotes the total number of generation steps.
The \textit{Tool Verifier} ensures correctness through syntax checking, execution testing, and domain validation:
\begin{equation}
\label{eq:validation}
P(\text{valid} \mid T_{\text{proposed}}) = P_{\text{syntax}} \cdot P_{\text{exec}} \cdot P_{\text{domain}}.
\end{equation}
Only tools that pass all validation checks proceed to the refinement stage.

\subsection{Atomic Tool Refinement}
To ensure the library evolves with high-quality, reusable assets, valid tools undergo atomic refinement before registration. The \textit{Atomic Decomposer} first breaks complex generated tools into fundamental ``cell tools''. The decomposition process is formalized as:
\begin{equation}
\label{eq:decomposition}
\begin{aligned}
\{A_1, \ldots, A_k\} &= \text{Decompose}(T),
\end{aligned}
\end{equation}
which maximizes the expected reuse improvement $\mathbb{E}[R_{\text{atomic}}]$:
\begin{equation}
\label{eq:reuse-bound}
\mathbb{E}[R_{\text{atomic}}] \geq k \cdot \mathbb{E}[R(T)] \cdot p_{\text{partial}},
\end{equation}
where $R(\cdot)$ represents the reuse utility function, $k$ denotes the number of atomic components derived from the decomposition, and $p_{\text{partial}}$ denotes the probability that a future problem requires only a subset of functions.
Intuitively, monolithic tools suffer from rigidity. Decomposing $T$ into $k$ atomic units unlocks partial reusability, allowing future queries to invoke specific sub-functions ($p_{\text{partial}}$) independently, which flexibility ensures the decomposed set yields higher cumulative utility than the single rigid tool.

The \textit{Redundancy Checker} acts as a gatekeeper. New atomic functions $A_{new}$ are compared against the library:
\begin{equation}
\label{eq:dedup}
A_{new} \in L_{t+1} \Leftrightarrow \max_{A_i \in L_t} \text{sim}(A_{new}, A_i) < \tau.
\end{equation}
Concurrently, the curator of the \textit{Dynamic Tool Registry} maintains library efficiency by pruning low-usage tools when capacity $C$ is exceeded, ensuring the library remains compact and relevant:
\begin{equation}
\label{eq:optimize}
L_{t+1} = L_t \setminus \{A_i : u(A_i) < \theta_{min} \land |L_t| > C\},
\end{equation}
where $u(A_i)$ denotes the historical usage count of tool $A_i$, and $\theta_{min}$ is the minimum usage threshold.

\subsection{Runtime Execution Engine}
Once the required tools are successfully retrieved or generated, the \emph{Tool Executor} integrates them into the final reasoning process. The solution synthesis is formalized as $S = \text{Solve}(P, L_t)$.
The whole framework closes the loop, applying the evolved capabilities of the library to synthesize the final answer $S$ for the user query.

\section{The SciEvo Benchmark}

\begin{figure}[t]
  \centering
  \includegraphics[width=\columnwidth]{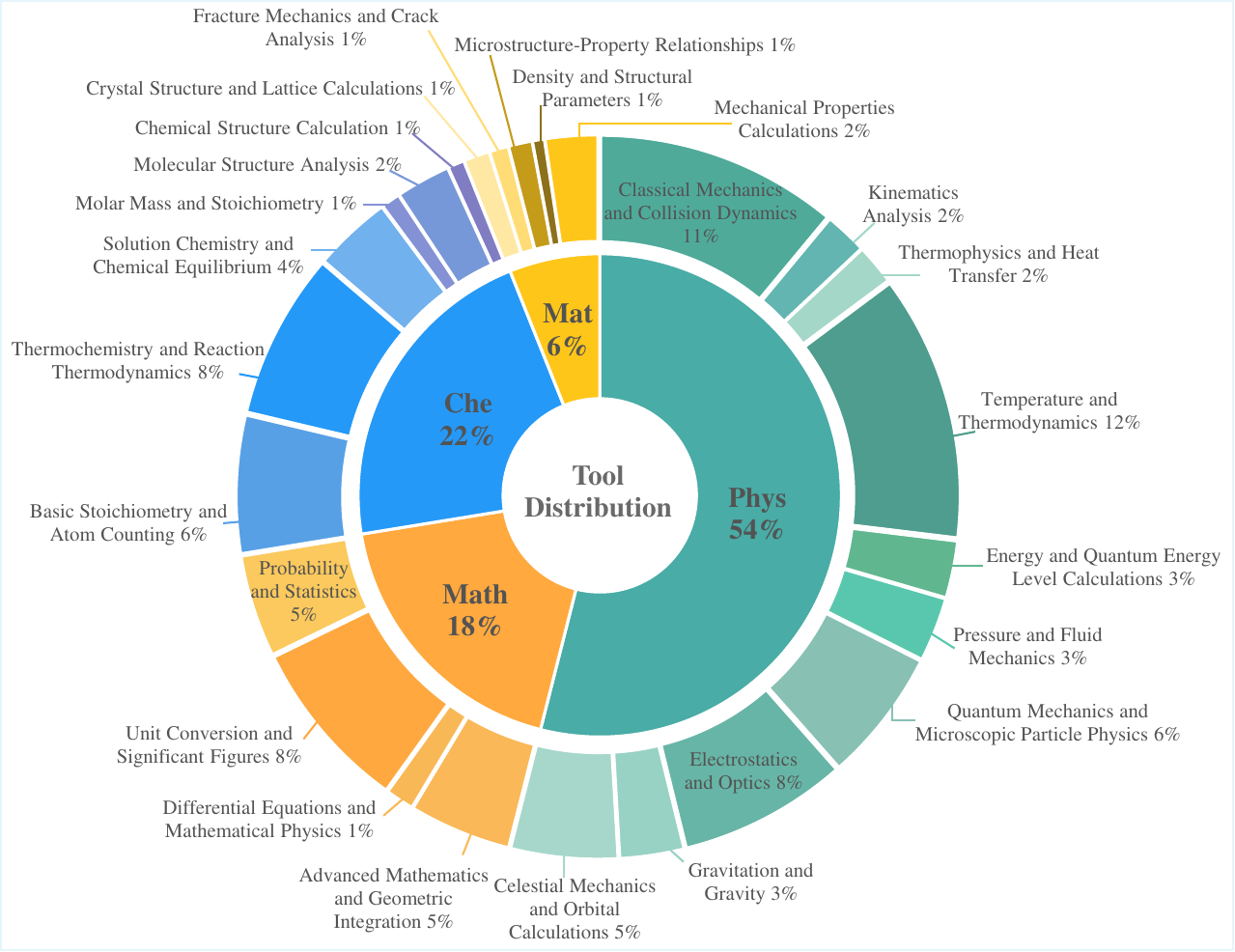}
  \caption{Tool distribution of the curated SciEvo benchmark. SciEvo covers 25 sub-disciplines across four major scientific fields: Physics (499 tools), Chemistry (192), Mathematics (171), and Materials (63), demonstrating comprehensive coverage of diverse scientific computational needs.}
  \label{fig:distribution}
\end{figure}

\subsection{Benchmark Construction}
A defining characteristic of SciEvo is its evolutionary construction paradigm. Unlike libraries curated from static codebases, tools from SciEvo are bootstrapped from scratch using the TTE framework, ensuring that every tool is pragmatically generated to address authentic scientific reasoning needs.

\paragraph{Seed Data Source.}
To construct a robust and diverse seed environment, we integrate high-quality scientific inquiries from three distinct sources, including SciEval \citep{sun2024scieval}, SciBench \citep{wang2023scibench}, and a proprietary materials science dataset focused on specialized domain calculations. We explicitly filter for computational problems that require multi-step reasoning and precise numerical solutions, filtering out purely knowledge-retrieval queries. To ensure the selected questions cover a comprehensive spectrum of scientific scenarios, we employ a semantic clustering-based stratified sampling strategy. Specifically, we embed all candidate questions using the embedding model \citep{reimers2019sentence} and perform K-Means clustering, subsequently sampling instances uniformly from each cluster to maximize problem diversity within the seed set. These pairs provide the problem contexts ($Q$) and ground-truth validation signals ($A$) required for reliable tool verification.

\paragraph{Tool Library Synthesis.}
We utilize the TTE-Zero framework to bootstrap the SciEvo tool library. By initializing the agent with an empty tool library and sequentially exposing it to the seed questions, we simulated a ``Tabula Rasa'' learning process. The agent generate, execute, and validate Python functions dynamically. Only atomic functions that successfully contributed to deriving the correct ground-truth answers are permanently inducted into the repository. The whole process yielded a verified library of 925 atomic tools, ensuring 100\% alignment between the toolset and the problem space.

\subsection{Taxonomy and Statistics}
To facilitate fine-grained analysis, we organize the synthesized tools into a hierarchical taxonomy using a hybrid classification strategy. 

\paragraph{Domain Classification.}
We apply Principal Component Analysis (PCA) on the vector embeddings of the generated tool descriptions to identify latent semantic clusters. These clusters are subsequently reviewed and refined by PhD-level domain experts to establish a precise taxonomy comprising 25 sub-disciplines across Physics (10), Chemistry (6), Materials Science (5), and Mathematics (4) as shown in Figure~\ref{fig:distribution}, which ensures the classification captures both computational semantics and canonical scientific distinctions.

\paragraph{Data Distribution.}
The complete SciEvo benchmark encompasses 1,590 evaluation instances supported by a library of 925 evolved tools. The domain-specific tool distribution spans four primary disciplines: Physics contains the largest subset with 499 tools, followed by Chemistry (192 tools), Mathematics (171 tools), and Materials (63 tools). As illustrated in Figure~\ref{fig:distribution}, the diverse composition ensures robust coverage of scientific computational primitives.

\subsection{Evaluation Metrics}
To simulate realistic resource constraints, all evaluations are conducted with a maximum tool library capacity of $C=500$. Under this setting, we assess performance using Accuracy (Acc), following standard protocols \citep{wang2023scibench,sun2024scieval}.
To quantify the utility and generalizability of the evolved library $\mathcal{T}$, we additionally define Tool Reuse Rate ($\text{TRR}@k$) for TTE-Zero as the proportion of tools that have been successfully reused at least $k$ times:
\begin{equation}
    \text{TRR}@k = \frac{|\{t \in \mathcal{T} \mid h(t) \ge k\}|}{|\mathcal{T}|},
\end{equation}
where $h(t)$ denote the hit-count for the tool $t$. We report $\text{TRR}@k$ at several increasing thresholds to capture different levels of utility, \textit{i.e.}, $\text{TRR}@1$ measures the fraction of non-redundant tools used at least once, while $\text{TRR}@5$ and $\text{TRR}@10$ identify the emergence of core scientific primitives.

For cross-domain evaluations, \textit{i.e.}, TTE-Adapt, we decompose the total tool library $\mathcal{T}$ into the pre-defined set $\mathcal{T}_{pre}$ and the newly evolved set $\mathcal{T}_{new}$. We introduce two stratified metrics to disentangle the sources of competence:
\begin{align}
    \text{TRR}_{evol}@k &= \frac{|\{t \in \mathcal{T}_{new} \mid h(t) \ge k\}|}{|\mathcal{T}_{new}|}, \label{eq:trr_evol} \\
    \text{TRR}_{trans}@k &= \frac{|\{t \in \mathcal{T}_{pre} \mid h(t) \ge k\}|}{|\mathcal{T}_{pre}|}, \label{eq:trr_trans}
\end{align}
where $\text{TRR}_{evol}@k$ serves as the primary benchmark metric for adaptation efficiency. A higher $\text{TRR}_{evol}$ indicates superior performance, signifying that the system has successfully consolidated novel domain knowledge into high-quality, reusable primitives rather than generating disposable scripts. Conversely, $\text{TRR}_{trans}$ monitors the substitution of prior knowledge. In cross-domain settings, a lower $\text{TRR}_{trans}$ is generally preferred as it reflects the mitigation of negative transfer, \textit{i.e.}, discarding irrelevant tools, provided it remains non-zero to ensure the retention of fundamental, domain-agnostic capabilities.

\begin{table*}[t]
\centering
\small
\begin{tabularx}{\textwidth}{l *{12}{>{\centering\arraybackslash}X}}
\toprule
\multirow{2}{*}{\textbf{Method}} &
\multicolumn{4}{c}{\textbf{SciBench}} &
\multicolumn{4}{c}{\textbf{SciEval}} &
\multicolumn{4}{c}{\textbf{SciEvo}} \\
\cmidrule(lr){2-5} \cmidrule(lr){6-9} \cmidrule(lr){10-13}
& \scriptsize TRR@1 & \scriptsize TRR@2 & \scriptsize TRR@5 & \scriptsize TRR@10 & 
  \scriptsize TRR@1 & \scriptsize TRR@2 & \scriptsize TRR@5 & \scriptsize TRR@10 & 
  \scriptsize TRR@1 & \scriptsize TRR@2 & \scriptsize TRR@5 & \scriptsize TRR@10 \\
\midrule
KTCE        & 0.17 & 0.10 & 0.05 & 0.02 & 0.06 & 0.04 & 0.03 & 0.02 & 0.31 & 0.20 & 0.09 & 0.04 \\
Creator     & 0.12 & 0.06 & 0.02 & 0.01 & 0.03 & 0.02 & 0.02 & 0.01 & 0.17 & 0.07 & 0.04 & 0.02 \\
CheMatAgent & 0.43 & 0.27 & 0.18 & 0.13 & 0.20 & 0.16 & 0.09 & 0.05 & 0.62 & 0.42 & 0.28 & 0.17 \\
\midrule
\rowcolor{gray!12}\textbf{TTE-Zero} &
\textbf{0.89} & \textbf{0.71} & \textbf{0.40} & \textbf{0.21} & 
\textbf{0.48} & \textbf{0.35} & \textbf{0.15} & \textbf{0.05} & 
\textbf{0.99} & \textbf{0.94} & \textbf{0.66} & \textbf{0.41} \\
\bottomrule
\end{tabularx}
\caption{Hierarchical analysis of tool reuse ($\text{TRR}@k$). We report the Tool Reuse Rate at thresholds $k=\{1, 2, 5, 10\}$. TTE-Zero achieves near-perfect utilization ($\text{TRR}@1 \approx 1.0$) on SciEvo and consistently maintains high reuse rates at stricter thresholds ($k=5, 10$), whereas baselines fail to generate high-frequency core primitives.}
\label{tab:tool-reuse}
\end{table*}

\begin{table}[t]
\centering
\small
\begin{tabularx}{\columnwidth}{l *{3}{>{\centering\arraybackslash}X}}
\toprule
\textbf{Method} & \textbf{SciBench} & \textbf{SciEval} & \textbf{SciEvo} \\
\midrule
Basic-COT       & 0.27 & 0.18 & 0.33 \\
Basic-POT       & 0.31 & 0.21 & 0.36 \\
Creator         & 0.27 & 0.22 & 0.49 \\
KTCE            & 0.37 & 0.24 & 0.55 \\
CheMatAgent  & 0.34 & 0.23 & 0.56 \\
\midrule
\rowcolor{gray!12}\textbf{TTE-Zero} &
\textbf{0.45} &
\textbf{0.30} &
\textbf{0.62} \\
\bottomrule
\end{tabularx}
\caption{Accuracy comparison across benchmarks. TTE-Zero consistently outperforms all baselines.}
\label{tab:scibench}
\end{table}

\section{Experiments}
\subsection{Experimental Setup}
\paragraph{Datasets.}
We evaluate our framework on three distinct benchmarks to assess both problem-solving accuracy and tool evolution efficiency, including SciBench \citep{wang2023scibench}, SciEval \citep{sun2024scieval}, and the curated SciEvo dataset. 

\paragraph{Baselines.}
We compare TTE-Zero against five representative baselines categorized into two paradigms. To evaluate fundamental reasoning capabilities without external tool support, we employ Basic-COT (Chain-of-Thought) and Basic-POT (Program-of-Thought). For agentic frameworks that utilize tools, we compare against Creator \citep{qian2023creator}, KTCE \citep{KTCE}, and CheMatAgent \citep{wu2025chemagent}. 
In the TTE-Adapt setting, we compare against a ``No Tool'' baseline and a ``Source Only'' baseline to isolate the performance gains attributed to domain-specific tool evolution.

\subsection{Implementation Details}
\label{subsec:implementation}
\paragraph{Model Architecture.}
We evaluate our framework using three LLMs, including GPT-4o, Qwen2.5-7B-Instruct, and GPT-3.5-turbo. Unless otherwise specified, the main experimental results are reported based on GPT-3.5-turbo with a sampling temperature of $0.3$ to balance diversity and determinism.

\paragraph{Retrieval and Ranking.}
We implement a dense retrieval pipeline using bge-m3 \citep{chen-etal-2024-m3} for embedding and bge-reranker-v2-m3 for re-ranking. For each sub-goal, the system retrieves the top-$k$ ($k=3$) relevant tools to provide focused context.

\paragraph{Tool Evolution and Deduplication.}
To maintain a compact and efficient library constrained to a maximum capacity of $C=500$, we employ strict semantic deduplication. We utilize CodeBERT \citep{feng-etal-2020-codebert} to compute semantic similarity between candidate tools and existing library entries. A new tool is strictly rejected if its maximum cosine similarity with any existing tool exceeds the threshold $\tau = 0.8$.

\paragraph{Evaluation Protocol.}
Final answer correctness is verified by a GPT-4.1-nano judge. We apply a relative tolerance of $10^{-5}$ for numerical results and require exact canonical matches for symbolic expressions. As for evaluation metrics, we report Accuracy (Acc) for solution correctness and the proposed Tool Reuse Rate ($\text{TRR}@k$, $\text{TRR}_{trans}@k$, and $\text{TRR}_{evol}@k$) to quantify the evolutionary quality of the tool library.

\begin{figure*}[t]
  \centering
  \includegraphics[width=\textwidth]{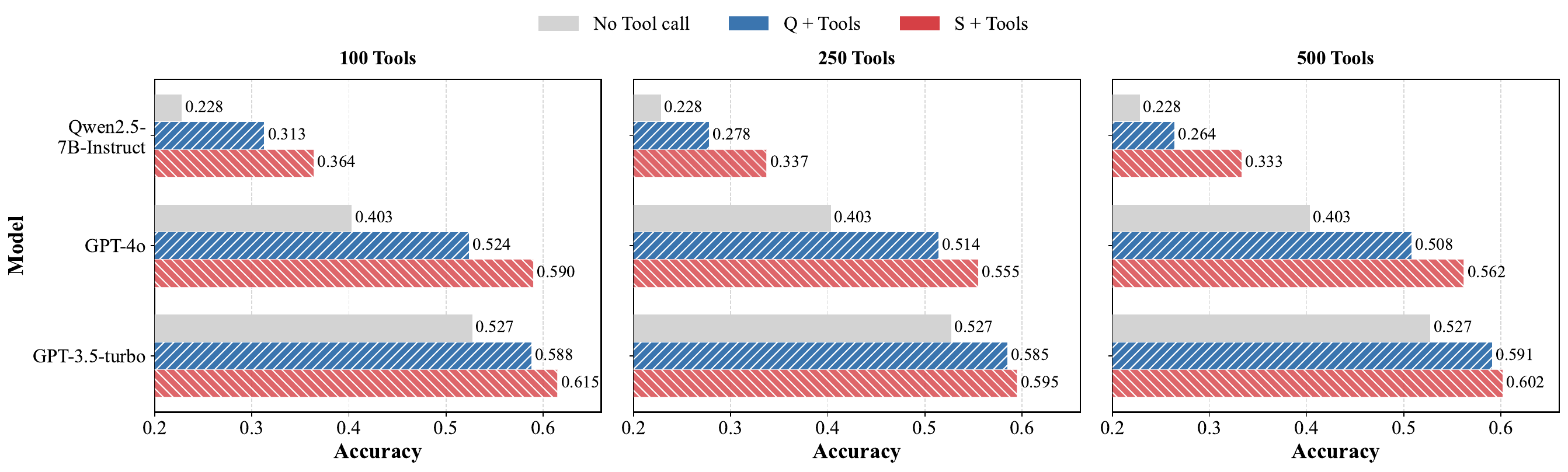}
  \caption{Accuracy comparison on SciEvo. We compare the ``No Tool call'' baseline against our TTE-Zero method using direct queries (``Q + Tools'') and Sub-goal Decomposition (``S + Tools'').}
  \label{fig:all_score_comparison_time}
\end{figure*}

\section{Results and Analysis}
\label{sec:results}

\subsection{Performance for TTE-Zero}
In this setting, the agent starts with an empty library ($L_0 = \emptyset$) to evaluate its capability to synthesize scientific primitives and solve real problems.
\paragraph{Comparative Analysis on Scientific Benchmarks.}
We first evaluate the performance of TTE-Zero against SOTA baselines across three benchmarks. As shown in Table \ref{tab:scibench}, TTE-Zero consistently establishes a new SOTA performance.
On the SciBench dataset, TTE-Zero achieves an accuracy of 0.45, significantly surpassing the strongest baseline KTCE (0.37) and the domain-specific CheMatAgent (0.34). The performance advantage is further amplified on the proposed SciEvo benchmark, where TTE-Zero reaches 0.62 accuracy compared to 0.56 for CheMatAgent and 0.55 for KTCE. The results demonstrate that evolving tools at test time provides a distinct advantage over static or retrieval-based paradigms, particularly for complex scientific problems requiring multi-step reasoning. Notably, TTE-Zero outperforms standard prompting strategies (Basic-COT and Basic-POT) by a wide margin, \textit{e.g.}, +0.29 improvement over Basic-COT on SciEvo, validating the necessity of external tool support.

\begin{table*}[t]
\centering
\resizebox{\textwidth}{!}{%
\begin{tabular}{l|c|cccc|cccc|c|cccc|cccc}
\toprule
\multirow{3}{*}{\textbf{Method}} & \multicolumn{9}{c|}{\textbf{Adaptation: Materials $\to$ Chemistry}} & \multicolumn{9}{c}{\textbf{Adaptation: Materials $\to$ Physics}} \\
\cmidrule(lr){2-10} \cmidrule(lr){11-19}
 & \multirow{2}{*}{\textbf{Acc} $\uparrow$} & \multicolumn{4}{c|}{\textbf{TRR}$_{trans}$ ($\downarrow$)} & \multicolumn{4}{c|}{\textbf{TRR}$_{evol}$ ($\uparrow$)} & \multirow{2}{*}{\textbf{Acc} $\uparrow$} & \multicolumn{4}{c|}{\textbf{TRR}$_{trans}$ ($\downarrow$)} & \multicolumn{4}{c}{\textbf{TRR}$_{evol}$ ($\uparrow$)} \\
 & & 1 & 2 & 5 & 10 & 1 & 2 & 5 & 10 & & 1 & 2 & 5 & 10 & 1 & 2 & 5 & 10 \\
\midrule
No Tool & 0.535 & - & - & - & - & - & - & - & - & 0.535 & - & - & - & - & - & - & - & - \\
Source Only & 0.561 & 0.26 & 0.13 & 0.04 & 0.02 & - & - & - & - & 0.585 & 0.38 & 0.20 & 0.03 & 0.01 & - & - & - & - \\
\rowcolor{gray!12} \textbf{TTE-Adapt} & \textbf{0.595} & \textbf{0.23} & \textbf{0.10} & \textbf{0.01} & \textbf{0.00} & \textbf{0.24} & \textbf{0.11} & \textbf{0.02} & \textbf{0.01} & \textbf{0.618} & \textbf{0.25} & \textbf{0.11} & \textbf{0.01} & \textbf{0.01} & \textbf{0.32} & \textbf{0.16} & \textbf{0.03} & \textbf{0.01} \\
\bottomrule
\end{tabular}%
}\caption{Performance on cross-domain adaptation (Source: Materials). We report Accuracy and Tool Reuse Rates (TRR) at $k \in \{1, 2, 5, 10\}$. $\text{TRR}_{trans}$ tracks retained source tools (lower is preferred to mitigate negative transfer), while $\text{TRR}_{evol}$ tracks new target tools (higher is better for knowledge consolidation).}
\label{tab:cross_domain_full}
\end{table*}

\paragraph{Analysis of Tool Evolution Quality.}
To understand whether the performance gain stems from efficient tool utilization or mere brute-force generation, we analyze the Tool Reuse Rate (TRR). Table \ref{tab:tool-reuse} presents the hierarchical reuse statistics.
A critical observation is the near-perfect utilization rate of TTE-Zero on the SciEvo dataset, achieving a TRR@1 of 0.99, which indicates that almost every generated tool was successfully reused to solve the target problem, minimizing computational waste. In contrast, baselines such as Creator (TRR@1 = 0.17) and KTCE (TRR@1 = 0.31) exhibit severe redundancy, where a vast majority of offline-generated tools are never used. Furthermore, TTE-Zero demonstrates superior capability in consolidating ``scientific primitives''. At the stricter threshold of $k=10$, it maintains a reuse rate of 0.41 on SciEvo and 0.21 on SciBench, whereas Creator drops to near zero (0.02 and 0.01, respectively), which confirms that TTE-Zero does not simply flood the library but actively evolves high-utility, reusable tools.

\paragraph{Ablation Study.}
We investigate the contribution of the sub-goal decomposition module by comparing two TTE variants: ``Q+Tools'' (using the original query) and ``S+Tools'' (using sub-goal decomposition) against the ``No Tool call'' baseline.
As illustrated in Figure \ref{fig:all_score_comparison_time}, both tool-augmented settings outperform the ``No Tool call'' baseline across all evaluated models, including Qwen2.5-7B, GPT-4o, GPT-3.5-turbo. Crucially, the ``S+Tools'' strategy consistently yields the highest accuracy. For instance, with a library size of 100 on Qwen2.5-7B, ``S+Tools'' achieves clear gains over ``Q+Tools'' (0.364 vs 0.313), which validates that breaking down complex scientific queries into granular sub-goals is essential for precise tool retrieval and execution, thereby maximizing the efficacy of the evolved tool library.

\subsection{Performance for TTE-Adapt}
We assess the plasticity of the TTE-Adapt framework by initializing it with a pre-defined tool library (\textit{e.g.}, Materials) and adapting it to novel target domains (\textit{e.g.}, Chemistry and Physics).

\paragraph{Cross-Domain Adaptation.} Table~\ref{tab:cross_domain_full} presents the adaptation performance. TTE-Adapt consistently outperforms the ``No Tool'' and ``Source Only'' baselines, achieving obvious accuracy gains in both settings. The performance improvement is driven by an adaptive substitution mechanism, where the system effectively mitigates negative transfer by pruning irrelevant source tools, \textit{i.e.}, reducing $\text{TRR}_{trans}@1$ from $0.26$ to $0.23$ in Chemistry, while simultaneously consolidating new knowledge into reusable primitives, which is evidenced by the substantial contribution of evolved tools ($\text{TRR}_{evol}@1 = 0.24$ in Chemistry and $0.32$ in Physics). The dynamic adjustment process confirms that TTE-Adapt successfully reshapes the tool distribution to align with the specific reasoning patterns of the target domain.

\section{Conclusion}
In this work, we identify and address the fundamental limitations of static tool paradigms in scientific reasoning. By introducing Test-Time Tool Evolution (TTE), we shift the role of LLM agents from passive tool selectors to active tool creators. TTE empowers agents to synthesize, verify, and evolve computational primitives during inference, ensuring that the tool space remains intrinsically isomorphic to the unbounded scientific problem space. Our extensive evaluations confirm that TTE not only establishes a new SOTA in reasoning accuracy but also enables robust tool adaptation across diverse domains. We believe that equipping agents with the capacity for autonomous tool evolution is a prerequisite for realizing the next generation of general-purpose scientific AI.

\section{Limitations}
While Test-Time Tool Evolution (TTE) introduces a promising paradigm for scientific reasoning, we acknowledge several limitations inherent to our current framework.

\paragraph{Inference Latency and Computational Cost.}
Unlike static retrieval-based methods, TTE requires synthesizing and verifying tools during inference. The dynamic evolution process inevitably incurs higher computational overhead and increased latency compared to simple tool selection. Future work could investigate lightweight ``meta-models'' to predict tool necessity, thereby skipping evolution for trivial queries.

\paragraph{Dependency on Base LLM Coding Capability.}
The efficacy of TTE is intrinsically bounded by the code generation capability of the backbone LLM. Our experiments demonstrate SOTA performance using the high-capacity models. However, performance degradation is observed with smaller, less capable open-source models (\textit{e.g.}, <7B parameters) that struggle with generating syntactically correct Python primitives. 

\paragraph{Safety and Sandboxing in Open-Ended Evolution.}
Allowing an agent to generate and execute arbitrary code at test time introduces potential safety risks, particularly in open-ended scientific exploration where generated scripts might inadvertently consume excessive resources or attempt unsafe operations. While our experiments are conducted in a strictly sandboxed environment with timeout constraints, scaling TTE to autonomous real-world systems will require more robust, semantic-level safety verification protocols beyond simple syntactic checks.

\section{Ethical Statement}
\label{sec:ethics}
We recognize that the dynamic generation of scientific tools introduces potential dual-use risks, particularly in domains such as chemistry or materials science where code could be misused for harmful applications (\textit{e.g.}, toxin synthesis). To mitigate these risks, we conducted a rigorous manual review of the entire evolved tool library prior to its public release. We strictly adhere to responsible disclosure practices and ensure that no tools enabling harmful applications are included in the released artifacts.

Regarding the artifacts released with this work, \textit{i.e.}, SciEvo benchmark, we confirm that all data sources are from public scientific repositories and are consistent with their intended research use. We have conducted a screening process to ensure no personally identifiable information (PII) or offensive content is included. The code and data are released under the MIT License to promote open research while restricting malicious use. We acknowledge that the system may reflect biases present in the underlying LLMs and scientific literature, and users should verify critical calculations before commercial deployment.
Finally, we acknowledge the use of AI assistants (\textit{e.g.}, ChatGPT) solely for linguistic polishing. All scientific claims, experimental designs, and data analyses remain our original work.

\bibliography{custom}

\clearpage

\appendix
\section{Complete Algorithmic Workflow}
\label{app:algorithm}
The appendix provides the end-to-end algorithmic details of Test-Time Tool Evolution (TTE) that are omitted from the main paper due to space constraints, including the full closed-loop evolution procedure, and the failure-handling logic that ensures robustness at test time.

\subsection{End-to-End Test-Time Tool Evolution}
\label{app:algorithm:full}

Algorithm~\ref{alg:tte_full} presents the full TTE pipeline. The same procedure covers both TTE-Zero (starting from an empty library) and TTE-Adapt (starting from a pre-defined source library), since the evolution loop is identical except for the initialization of the registry.
For clarity and reproducibility, we explicitly distinguish:
(i) $\tau_{\textsf{ret}}$ for retrieval acceptance (whether to reuse a retrieved tool), and (ii) $\tau_{\textsf{dup}}$ for deduplication (whether a new atomic tool is considered redundant).
These two thresholds need not be identical and can be tuned independently.

\subsection{Failure Handling and Fallback}
\label{app:fallback}
TTE is designed to be robust under imperfect tool synthesis. When verification fails (syntax / runtime / domain constraints), the proposed tool is \emph{not} registered. Downstream execution can either attempt to solve the sub-goal via reasoning-only mode or proceed with partial tool chains when intermediate values are still available. In our implementation, we primarily use a conservative strategy: only verified tools are registered, and failed tool generation triggers a lightweight fallback to direct reasoning or Program-of-Thought, ensuring the system degrades gracefully rather than accumulating faulty tools.

\begin{algorithm}[t]
\caption{Complete Test-Time Tool Evolution.}
\label{alg:tte_full}
\small
\begin{algorithmic}[1]
\Require User problem $P$, initial tool library $L$, library capacity $C$
\Ensure Solution $S$ (or \textsc{Fail})
\State $\mathcal{O} \leftarrow \textsc{Decompose}(P)$
\State $\textsf{chain} \leftarrow [\ ]$ 
\For{each operation $O_i \in \mathcal{O}$}
    \State $\mathcal{T}_i \leftarrow \textsc{RetrieveTopK}(L, O_i, k)$ 
    \State $(T^\star, s^\star) \leftarrow \arg\max_{T \in \mathcal{T}_i} \textsc{sim}(T, O_i)$
    \If{$s^\star \ge \tau_{\textsf{ret}}$}
        \State $\textsf{chain}.\textsc{append}(T^\star)$; $u(T^\star)\mathrel{+}=1$
    \Else
        \State $T_{\textsf{new}} \leftarrow \textsc{SynthesizeTool}(P, O_i)$
        \State $\textsf{ok} \leftarrow \textsc{VerifyTool}(T_{\textsf{new}})$ 
        \If{\textsf{ok} = \textbf{false}}
            \State \textbf{continue} 
        \EndIf
        \State $\mathcal{A} \leftarrow \textsc{AtomicDecompose}(T_{\textsf{new}})$ 
        \For{each atomic tool $A \in \mathcal{A}$}
            \If{$\max_{T \in L} \textsc{sim}_{\textsf{dup}}(A, T) < \tau_{\textsf{dup}}$}
                \State $L \leftarrow L \cup \{A\}$; $u(A)\leftarrow 1$
            \Else
                \State $T_{\textsf{match}} \leftarrow \arg\max_{T \in L}\textsc{sim}_{\textsf{dup}}(A,T)$
                \State $u(T_{\textsf{match}})\mathrel{+}=1$ 
            \EndIf
        \EndFor
        \State $L \leftarrow \textsc{PruneIfNeeded}(L, C)$ 
        \State $\textsf{chain}.\textsc{append}(T_{\textsf{new}})$
    \EndIf
\EndFor
\State $S \leftarrow \textsc{ExecuteChain}(P, \textsf{chain})$
\If{$S=\textsc{Fail}$}
    \State $S \leftarrow \textsc{Fallback}(P)$ \
\EndIf
\State \Return $S$
\end{algorithmic}
\end{algorithm}

\section{Prompts for Each Agent Module}
\label{app:prompts}

This section details the system prompts designed for the three core modules of the TTE framework. We enforce strict JSON or XML-based output formats to ensure robust parsing and seamless integration with the Python execution environment. 
\subsection{Structured Task Decomposition}
\label{app:prompts:decompose}
The \textit{Problem Analyzer} translates high-level scientific queries into linear execution plans using the prompt in Figure \ref{fig:prompt_decompose}.

\begin{figure*}[h]
\centering
\begin{promptbox}[title=Prompt for Problem Analyzer]
[SYSTEM]
You are an expert computational scientist with broad interdisciplinary expertise.
You excel at decomposing complex scientific problems into a sequence of programmable computation steps that can be executed by predefined atomic tools.

[TASK]
Decompose the user problem into a list of concrete computational subtasks.
Each subtask must be a specific computation step (not analysis, discussion, or description).
The subtasks must form a coherent execution order: outputs of earlier steps can be used as inputs to later steps. Decompose until the full computation pipeline is covered, but do NOT provide the final numerical answer.

[STRICT OUTPUT FORMAT]
Return STRICT JSON only (no extra text, no Markdown, no comments):
{
  "original_problem": "...",
  "subtasks": [
    {"step": 1, "description": "..."},
    {"step": 2, "description": "..."},
    ...
  ]
}

[USER]
{query}
\end{promptbox}
\caption{The prompt used by the Problem Analyzer to decompose user queries into structured execution plans.}
\label{fig:prompt_decompose}
\end{figure*}

\subsection{Dynamic Tool Retrieval}
\label{app:prompts:toolcall}
The \textit{Tool Retriever} selects existing primitives from the library using the prompt in Figure \ref{fig:prompt_toolcall}, which enforces a "no-hallucination" policy.

\begin{figure*}[h]
\centering
\begin{promptbox}[title=Prompt for Tool Executor]
[SYSTEM]
You are a strict tool-calling agent. You MUST respond only by calling one predefined AtomicTool.
Your goal is to select the single most relevant tool from the provided tool catalog and call it.

[CORE RULES]
1) Do NOT answer the question directly. Do NOT do calculations in natural language.
2) Call exactly ONE tool. If no tool matches, return an empty tool-call list.
3) Do NOT invent tools. Only use tools provided by the system.
4) Output MUST be in standard OpenAI tool-call JSON format. No extra text.

[USER]
Problem: {sub_question_or_operation}
Tool catalog: {tool_catalog_with_signatures}
\end{promptbox}
\caption{The prompt used by the Tool Executor to invoke primitives from the library.}
\label{fig:prompt_toolcall}
\end{figure*}

\subsection{Tool Synthesis and Reasoning}
\label{app:prompts:synthesis}
Figure \ref{fig:prompt_synthesis} displays the hybrid prompt used by the generative tool synthesis module. It handles both tool synthesis and final answer generation.

\begin{figure*}[h]
\centering
\begin{promptbox}[title=Prompt for Tool Synthesizer and Tool Executor]
[SYSTEM]
You are an expert team composed of specialists from multiple scientific disciplines.
You can perform rigorous interdisciplinary computation by combining step-by-step sub-questions with executable Python code.

[INPUT]
Main question:
{main_question}

Sub-questions (process in order). Each item provides a sub-question and its corresponding code.
If the code is empty / null, treat it as "missing".
{(step_i, sub_question_i, code_i)}  # repeated for i=1..n

[EXECUTION RULES]
1) If a sub-question provides non-empty and valid code, simulate its execution and write down the structured result.
2) If a sub-question has missing code or the provided code cannot solve the sub-question, generate ONE runnable Python function:
   - snake_case function name
   - include a clear docstring with I/O specification and units
   - include a minimal test example
   Then simulate executing your function on the test example.

[NUMBERS AND UNITS]
- Use scientific notation with 6 significant digits for floating-point numbers.
- For probabilities/ratios in [0,1], round to 4 decimal places.
- All physical quantities must include SI units. If units are missing, explicitly state the default assumption.

[STRICT OUTPUT FORMAT]
Part 1: For missing-code sub-questions ONLY, output a JSON list wrapped in <code> ... </code>:
<code>
[
  {
    "sub_question": "...",
    "name": "function_name",
    "code": "full Python function as a string",
    "text_description": "brief function summary",
    "io_description": {"input": "...", "output": "..."},
    "test_example": {"input": {...}, "result": "... or null"},
    "error": "optional"
  }
]
</code>

Part 2: Output the final answer wrapped in <answer> ... </answer>:
<answer>
Plain natural-language conclusion (do not mention "code" or "function").
</answer>

[CONCISENESS]
No extra text is allowed outside <code> and <answer>.
\end{promptbox}
\vspace{-10pt}
\caption{The hybrid prompt used for synthesizing new tools and deriving the final scientific conclusion.}
\label{fig:prompt_synthesis}
\end{figure*}

\section{Subject-wise Results on SciEvo}
\label{app:subject_results}
Table~\ref{tab:main_results} reports subject-wise performance on SciEvo under three settings:
(i) ``No Tools'' (direct inference), (ii) ``Q+Tools'' (retrieve tools using the original question as query),
and (iii) ``S+Tools'' (retrieve tools using decomposed sub-questions as queries).


\begin{figure*}[t]
  \centering
  \includegraphics[width=\textwidth]{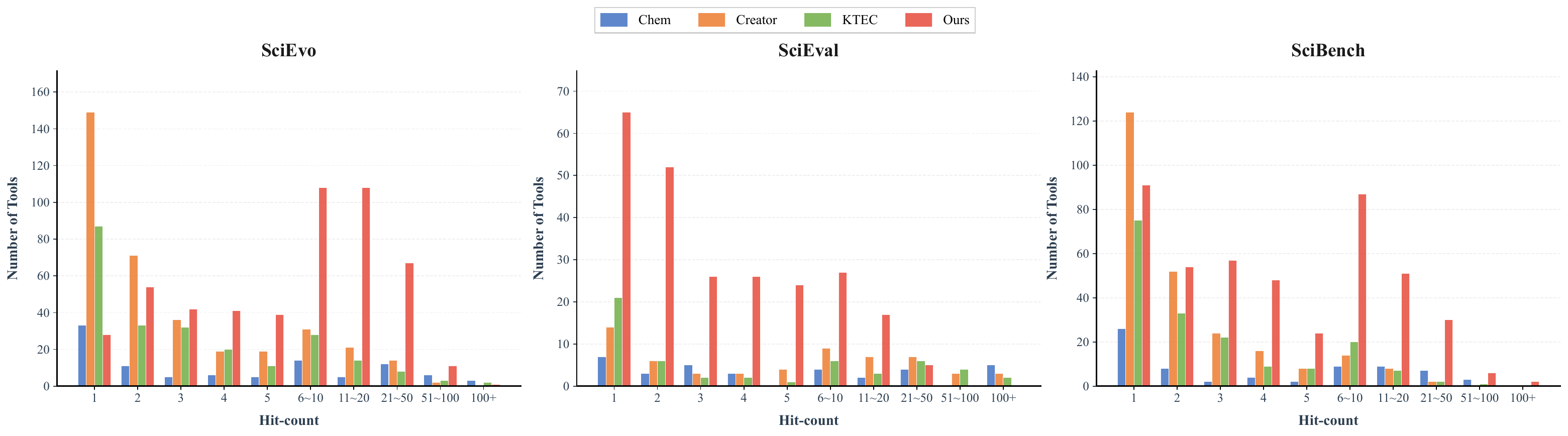}
  \caption{Histogram of tool usage frequency (Hit-count) across three benchmarks. The x-axis represents the reuse frequency of tools, and the y-axis denotes the number of tools.}
  \label{fig:tool_usage_distribution_bars_only}
\end{figure*}

\begin{figure*}[t]
  \centering
  \includegraphics[width=\textwidth]{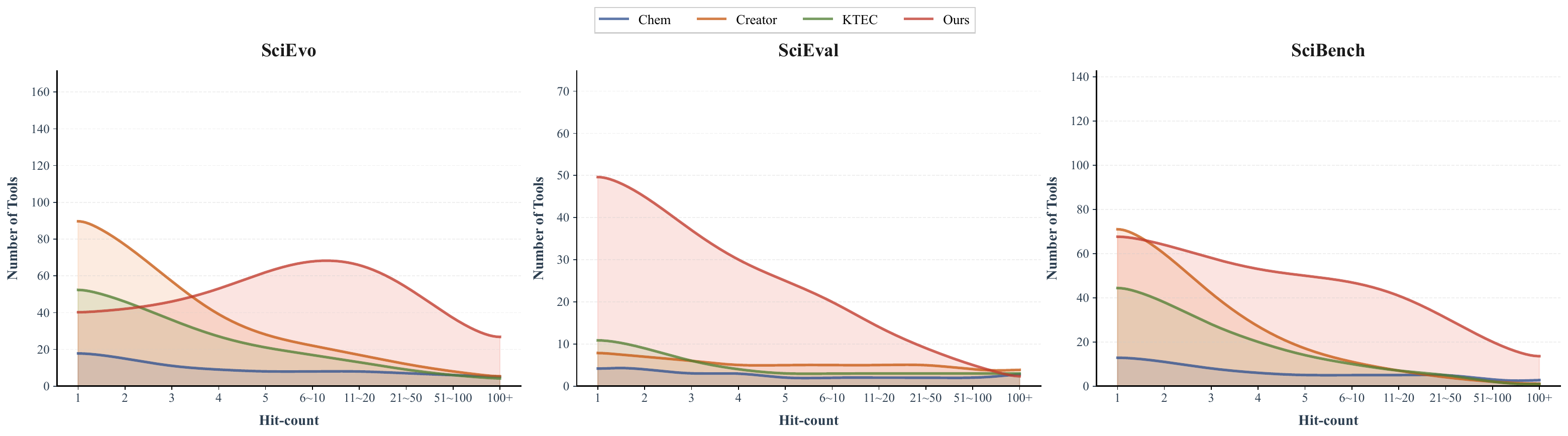}
  \caption{Kernel Density Estimation (KDE) of tool utilization rates. The distribution curves visualize the distributional shift in tool reusability.}
  \label{fig:tool_usage_distribution_curves_only}
\end{figure*}

\begin{table*}[t]
\centering
\small

\begin{tabularx}{\textwidth}{l l *{7}{>{\centering\arraybackslash}X}}
\toprule
\multirow{2}{*}{\textbf{Model}} & \multirow{2}{*}{\textbf{Subject}} & \multirow{2}{*}{\textbf{No Tools}} & 
\multicolumn{3}{c}{\textbf{Q + Tools}} & \multicolumn{3}{c}{\textbf{S + Tools}} \\
\cmidrule(lr){4-6} \cmidrule(lr){7-9}
& & & \textbf{100} & \textbf{250} & \textbf{500} & \textbf{100} & \textbf{250} & \textbf{500} \\
\midrule

\multirow{4}{*}{GPT-3.5-turbo} 
  & Che  & 0.548 & 0.580 & 0.540 & 0.551 & \textbf{0.647} & 0.589 & 0.619 \\
  & Math & 0.503 & 0.548 & 0.585 & \textbf{0.598} & 0.564 & 0.552 & 0.546 \\
  & Phy  & 0.524 & 0.600 & 0.629 & \textbf{0.642} & 0.634 & 0.609 & 0.639 \\
  & Mat  & 0.535 & 0.625 & 0.589 & 0.607 & 0.617 & \textbf{0.631} & 0.607 \\
\midrule

\multirow{4}{*}{GPT-4o} 
  & Che  & 0.412 & 0.536 & 0.539 & 0.539 & \textbf{0.638} & 0.594 & 0.578 \\
  & Math & 0.341 & 0.376 & 0.366 & 0.366 & \textbf{0.502} & 0.482 & 0.487 \\
  & Phy  & 0.346 & 0.533 & 0.534 & 0.487 & \textbf{0.615} & 0.557 & 0.576 \\
  & Mat  & 0.515 & \textbf{0.652} & 0.618 & 0.642 & 0.605 & 0.588 & 0.608 \\
\midrule

\multirow{4}{*}{Qwen2.5-7B} 
  & Che  & 0.211 & 0.317 & 0.291 & 0.312 & \textbf{0.400} & 0.358 & 0.352 \\
  & Math & 0.175 & 0.246 & 0.246 & 0.201 & 0.256 & 0.241 & \textbf{0.286} \\
  & Phy  & 0.227 & 0.332 & 0.302 & 0.304 & \textbf{0.368} & 0.365 & 0.360 \\
  & Mat  & 0.301 & 0.358 & 0.274 & 0.240 & \textbf{0.433} & 0.384 & 0.334 \\
\bottomrule
\end{tabularx}

\caption{Subject-wise performance on the SciEvo benchmark based on TTE framework. Che: Chemistry, Math: Mathematics, Phy: Physics, Mat: Materials. Sub-question decomposition (``S+Tools'') consistently outperforms main question input (``Q+Tools'') across all subjects.}
\label{tab:main_results}
\end{table*}


Across all model--subject pairs, tool augmentation provides clear benefits over direct inference. More importantly, sub-question driven retrieval (S+Tools) tends to be more robust and yields higher peak performance than retrieving tools using the original question (Q+Tools), especially in Chemistry and Physics, where tool selection is sensitive to units, constants, and domain-specific formulas.

A key pattern is that Chemistry exhibits the largest gain from structured decomposition: Chemistry queries often mix multiple operations (unit conversions, ideal gas relations, stoichiometry), where decomposed sub-questions provide a sharper semantic signal for retrieval and reduce the chance of selecting irrelevant tools. Materials science, in contrast, often has higher baseline performance and may require fewer distinct atomic operations per problem, resulting in relatively smaller incremental gains from decomposition.

We also observe that the best configuration depends on both the model and the subject. This motivates the library-size analysis in Appendix~\ref{app:tool_overload}, which explains why increasing the tool inventory does not always monotonically improve performance under question-level retrieval.

\section{Analysis of Tool Reusability}
\label{app:hitcount}
We investigate the reusability of generated tools by analyzing their invocation frequency (hit-count) across benchmarks. As visualized in the histograms (Figure~\ref{fig:tool_usage_distribution_bars_only}) and Kernel Density Estimation curves (Figure~\ref{fig:tool_usage_distribution_curves_only}), the baseline methods exhibit a severe left-skewed distribution, where the vast majority of tools are concentrated in the lowest frequency bins ($1\sim2$ uses). The heavy reliance on ``disposable'' tools suggests that baselines tend to overfit specific queries with monolithic scripts, resulting in high redundancy and poor transferability.

In contrast, TTE demonstrates a significant right-shift in probability density, effectively redistributing the mass towards moderate-to-high frequency ranges (\textit{e.g.}, $10\sim50+$). This phenomenon indicates a qualitative transition from generating ad-hoc solutions to discovering atomic computational primitives. By evolving tools that capture fundamental operations (\textit{e.g.}, canonical formulas or unit conversions), TTE reduces redundancy and ensures that the learned tool library consists of generalized modules capable of solving diverse scientific problems through composition.

\section{Explanation of Evaluation Metrics}
\label{app:metrics}
\subsection{Metrics for TTE-Zero}
In the TTE-Zero setting, where the system evolves a tool library $\mathcal{T}$ entirely from scratch, we employ the Tool Reuse Rate ($\mathrm{TRR}@k$) to quantify the utility and generalizability of the synthesized functions.
\begin{equation}
    \mathrm{TRR}@k = \frac{|\{t \in \mathcal{T} \mid h(t) \ge k\}|}{|\mathcal{T}|}.
\end{equation}
We interpret $\mathrm{TRR}@k$ across increasing thresholds to capture distinct dimensions of evolutionary quality.
$\mathrm{TRR}@1$ measures the fraction of non-redundant tools that are successfully executed at least once. A value approaching $1.0$ indicates minimal computational waste, signifying that the generation process is precise and avoids creating ``dead code'' or hallucinated functions.
$\mathrm{TRR}@2$ reflects immediate transferability, identifying tools that are robust enough to address multiple distinct queries rather than overfitting a single instance.
Crucially, higher-order metrics like $\mathrm{TRR}@5$ and $\mathrm{TRR}@10$ serve as indicators for the emergence of core scientific primitives. A high value at these thresholds suggests that the system has autonomously discovered and consolidated fundamental domain operators (\textit{e.g.}, specific unit converters or thermodynamic equation solvers) that are essential for solving a broad class of problems.

\subsection{Metrics for TTE-Adapt}
\label{app:trr_adapt}
In the cross-domain adaptation setting, \textit{i.e.}, TTE-Adapt, the system must balance stability (retaining useful prior knowledge) and plasticity (acquiring new domain-specific capabilities). To rigorously evaluate this dynamic, we decompose the final tool library $\mathcal{T}$ into two disjoint sets:
(i) the pre-defined source subset $\mathcal{T}_{\textsf{pre}}$, transferred from the source domain, and
(ii) the newly evolved subset $\mathcal{T}_{\textsf{new}}$, synthesized autonomously during target-domain inference.
We introduce two stratified metrics $\mathrm{TRR}_{\mathrm{trans}}@k$ and $\mathrm{TRR}_{\mathrm{evol}}@k$ to disentangle the sources of competence.

\paragraph{$\mathbf{TRR}_{\mathbf{evol}}@k$.} Metric for Knowledge Consolidation (Higher is Better). This is the primary metric for evaluating the quality of adaptation. In standard code generation approaches, models often generate ``disposable'' scripts, \textit{i.e.}, one-off solutions that solve a single query but lack generalizability. A high $\mathrm{TRR}_{\mathrm{evol}}@k$ (especially for $k \ge 5$) indicates that the system has successfully distilled the ``physical laws'' or ``core primitives'' of the new domain into reusable atomic functions. It confirms that the library growth is efficient: the system solves many problems with a compact set of high-quality new tools, rather than overfitting with a bloated library of redundant scripts.

\paragraph{$\mathbf{TRR}_{\mathbf{trans}}@k$.} Metric for Negative Transfer Mitigation. This metric monitors the utility of prior knowledge. In cross-domain settings (\textit{e.g.}, Materials $\to$ Chemistry), we expect $\mathrm{TRR}_{\mathrm{trans}}$ to decrease compared to in-domain settings. A lower value implies the system correctly identifies and prunes source tools that are irrelevant or harmful to the target domain (\textit{e.g.}, discarding a specific material property calculator that is invalid for molecules). However, this value should remain non-zero. A non-zero retention rate signifies the preservation of domain-agnostic capabilities (\textit{e.g.}, basic algebra, statistical functions, unit conversion) that are universally applicable.

\paragraph{The Substitution Effect.}
By analyzing the joint trajectory of $(\mathrm{TRR}_{\mathrm{trans}}, \mathrm{TRR}_{\mathrm{evol}})$, we can diagnose the adaptation strategy. Our empirical results shown in Table ~\ref{tab:cross_domain_full} demonstrate a Substitution Effect: as the domain gap increases, TTE autonomously reduces its reliance on $\mathcal{T}_{\textsf{pre}}$ (lower $\mathrm{TRR}_{\mathrm{trans}}$) and compensates by up-regulating the synthesis of $\mathcal{T}_{\textsf{new}}$ (higher $\mathrm{TRR}_{\mathrm{evol}}$). This contrasts with static baselines that suffer from ``forced fit'', attempting to solve new problems with mismatched old tools.

\section{The Tool Overload Phenomenon}
\label{app:tool_overload}
The empirical analysis of Table~\ref{tab:main_results} reveals a counter-intuitive non-monotonic trend: expanding the tool inventory from 100 to 500 atomic primitives does not consistently translate to performance gains. In specific configurations, particularly those relying on direct query-to-tool matching, increasing the library size paradoxically degrades problem-solving accuracy. We term this observation the ``Tool Overload Phenomenon''.

\paragraph{Theoretical Analysis.}
\label{app:tool_overload:hypothesis}
We attribute this degradation to the inherent tension between library richness and retrieval robustness. As the tool library expands, the semantic density of the vector space increases, inevitably shrinking the distance between the optimal tool and high-similarity distractors. This leads to retrieval collisions, where semantically adjacent but functionally distinct tools, \textit{e.g.}, two variations of a thermodynamic calculator with slightly different input assumptions, crowd out the correct candidate during nearest-neighbor search.

Furthermore, even when the correct tool is successfully retrieved, the presence of these high-similarity distractors within the context window introduces significant contextual interference. The language model is forced to perform fine-grained discrimination among subtly different function signatures, which increases the cognitive load of the selection process. The ``choice paralysis'' consumes the model's reasoning capacity, increasing the likelihood of selecting a suboptimal tool or hallucinating parameters, thereby neutralizing the theoretical benefits of a larger capability set.

\paragraph{Implications for Scalable Agent Systems.}
These findings suggest that scaling tool libraries requires more than simply accumulating functions. It demands architectural innovations in the retrieval mechanism. To mitigate the noise introduced by library expansion, future systems must move beyond flat similarity search. Potential solutions include hierarchical indexing strategies that first isolate the relevant tool domain before selecting specific atomic functions, or uncertainty-aware retrieval mechanisms that dynamically adjust the number of retrieved candidates based on the semantic ambiguity of the query.

\section{Case Studies}
\label{app:case_study}

We provide a detailed examination of two scientific reasoning scenarios to illustrate the specific failure modes of static tool libraries and how the Test-Time Tool Evolution (TTE) framework resolves them.

\begin{table*}[t]
\centering
\small
\begin{tabularx}{\textwidth}{p{0.17\textwidth} X}
\toprule
\textbf{Setting} & \textbf{Outcome and Diagnosis} \\
\midrule
No Tools &
Predicted $76.9~\mathrm{g}~\mathrm{mol}^{-1}$ (incorrect). The model likely applies an incorrect rearrangement or loses unit consistency without executable validation. \\
\midrule
Q+Tools &
Predicted $1.73\times 10^{2}~\mathrm{g}~\mathrm{mol}^{-1}$ (incorrect).
However, the tool context may contain many irrelevant functions, increasing selection noise and making the success less reliable across instances. \\
\midrule
S+Tools &
Predicted $1.69\times 10^{2}~\mathrm{g}~\mathrm{mol}^{-1}$ (correct).
Decomposition isolates the missing operation ``compute molar volume via $V_m = RT/P$'' and triggers targeted tool synthesis, producing an explicit executable function with unit handling. \\
\bottomrule
\end{tabularx}
\caption{Execution results for Case 1. Sub-question decomposition enables targeted tool synthesis and reduces retrieval ambiguity.}
\label{tab:case_molar_mass}
\end{table*}

\begin{table*}[h]
\centering
\resizebox{\textwidth}{!}{%
\begin{tabular}{c l l l l}
\toprule
\textbf{Step} & \textbf{Sub-Goal Description} & \textbf{Tool Action} & \textbf{Tool Status} & \textbf{Intermediate Result} \\
\midrule
1 & Convert density from $kg/m^3$ to $g/L$ & \texttt{convert\_density(1.23)} & \textcolor{blue}{Retrieved} & $1.23~g/L$ \\
2 & Convert pressure from $kPa$ to $Pa$ & \texttt{convert\_pressure(20)} & \textcolor{blue}{Retrieved} & $20,000~Pa$ \\
\rowcolor{gray!12} 3 & Calculate molar volume $V_m$ ($PV=nRT$) & \texttt{calculate\_molar\_volume(20000, 330)} & \textcolor{red}{\textbf{Evolved}} & $13.738~L/mol$ \\
4 & Calculate molar mass $M = \rho \times V_m$ & \texttt{calculate\_molar\_mass(1.23, 13.738)} & \textcolor{blue}{Retrieved} & $\mathbf{169.0}~g/mol$ \\
\bottomrule
\end{tabular}%
}
\caption{Step-by-step execution trace for Case 1. The \textit{Tool Status} column highlights the system's adaptive behavior: it retrieves existing tools for standard operations (Steps 1, 2, 4) but autonomously evolves a new tool for the missing primitive in Step 3.}
\label{tab:case1_trace}
\end{table*}

\begin{figure*}[t]
\centering
\begin{mycodebox}
def calculate_molar_volume(pressure_pa, temperature_k):
    """
    Compute molar volume Vm under the ideal gas law: Vm = RT/P.

    Args:
        pressure_pa (float): pressure in Pa
        temperature_k (float): temperature in K

    Returns:
        float: molar volume in L/mol
    """
    R = 8.314462618  # J/(mol*K)
    vm_m3_per_mol = (R * temperature_k) / pressure_pa
    vm_L_per_mol = vm_m3_per_mol * 1000.0
    return vm_L_per_mol
\end{mycodebox}
\caption{Excerpt of a synthesized atomic function for Case 1 that enables correct molar mass computation.}
\label{fig:case_molar_volume_code}
\end{figure*}

\subsection{Case 1: Molar Mass Estimation}
\label{app:case_study:molar_mass}

\paragraph{Problem Definition.}
The task requires estimating the molar mass of a gaseous compound given its density ($1.23~\mathrm{kg}~\mathrm{m}^{-3}$), temperature ($330~\mathrm{K}$), and pressure ($20~\mathrm{kPa}$). The ground truth is $169~\mathrm{g}~\mathrm{mol}^{-1}$. This problem serves as a critical test of the system's ability to handle multi-step reasoning, strict unit consistency, and missing computational primitives.

\paragraph{Performance Comparison.}
As summarized in Table~\ref{tab:case_molar_mass}, baseline methods struggle with either hallucination or precision loss. The \textit{No Tools} baseline relies on parametric knowledge, resulting in a physically plausible but numerically incorrect value ($76.9~\mathrm{g}~\mathrm{mol}^{-1}$), likely due to a misapplication of the Ideal Gas Law rearrangement. The \textit{Q+Tools} setting retrieves a generic density calculator but suffers from noise in the context window, yielding an approximate result ($173~\mathrm{g}~\mathrm{mol}^{-1}$). In contrast, our \textit{S+Tools} framework achieves the exact analytical solution ($169~\mathrm{g}~\mathrm{mol}^{-1}$) by enforcing a structured execution path.

\paragraph{Evolutionary Execution Trace.}
The core advantage of TTE is visualized in Table~\ref{tab:case1_trace}, which details the step-by-step resolution process. The \textit{Problem Analyzer} first decomposes the complex query into four atomic sub-questions. 
For Steps 1, 2, and 4, the system successfully identifies high-similarity matches in the existing library and retrieves the standard unit conversion and arithmetic tools. 
However, at Step 3, the system encounters a gap: the library contains generic gas law functions but lacks a specific primitive to calculate molar volume directly from pressure and temperature. Detecting this retrieval failure, the \textit{Tool Synthesizer} is triggered. It generates a dedicated atomic function \texttt{calculate\_molar\_volume} (shown in Figure~\ref{fig:case_molar_volume_code}), which correctly handles the gas constant $R$ and unit conversion from $m^3$ to $L$. The ``evolved'' tool is then immediately executed, bridging the computational gap that caused baselines to fail.

\subsection{Case 2: Electroplating Stoichiometry}
\label{app:case_study:electroplating}

\paragraph{Problem Definition.}
The problem involves calculating the mass of silver deposited on a tray via electrolysis (current: $8.46~\mathrm{A}$, time: $8.0~\mathrm{h}$) and subsequently determining the tray's surface area given a plating thickness of $0.00254~\mathrm{cm}$ and density of $10.5~\mathrm{g/cm^3}$. The ground truth values are $33.98~\mathrm{g}$ for mass and $1275.6~\mathrm{cm^2}$ for area. This task requires chaining Faraday's laws of electrolysis with geometric volume-area relationships.

\paragraph{Performance Comparison.}
Table~\ref{tab:case_electroplating} contrasts the outcomes. The \textit{No Tools} baseline typically fails to integrate the physics constants (Faraday's constant) correctly, leading to magnitude errors. The \textit{Q+Tools} model generates a monolithic function that correctly identifies the physics formulas but likely misinterprets the time parameter or stoichiometry context, yielding a result of $285.6~\mathrm{g}$, which deviates significantly from the ground truth. In contrast, \textit{S+Tools} achieves high precision ($31.6~\mathrm{g}$ and $1283~\mathrm{cm^2}$) by decomposing the problem into charge calculation, stoichiometric conversion, and geometric derivation, validating each step independently.

\paragraph{Evolutionary Execution Trace.}
Table~\ref{tab:case2_trace} illustrates the adaptive workflow. The system decomposes the physics problem into sequential logic.
Steps 1 and 4 utilize existing library tools for basic unit conversion and mass-mole relations.
However, for Step 2 (calculating moles of electrons from charge), Step 5 (calculating volume), and Step 6 (deriving area from volume and thickness), the retrieval system returned low-similarity matches. Consequently, the \textit{Tool Synthesizer} evolved dedicated primitives: \texttt{calculate\_moles\_of\_electrons} (incorporating Faraday's constant) and \texttt{calculate\_area}. Figure~\ref{fig:case_area_code} displays the evolved area calculation tool, which explicitly handles the geometric relationship $A = V/t$, bridging the gap between chemical output and geometric input.

\begin{table*}[t]
\centering
\small
\begin{tabularx}{\textwidth}{p{0.17\textwidth} X}
\toprule
\textbf{Setting} & \textbf{Outcome and Diagnosis} \\
\midrule
No Tools &
Predicted incorrect values due to lack of domain constants (Faraday's constant) and geometric formulas. Hallucination of intermediate values is common. \\
\midrule
Q+Tools &
Predicted $285.6~\mathrm{g}$ and $10,800~\mathrm{cm^2}$ (incorrect).
The model attempted a single-step calculation. While the code logic was syntactically correct, the monolithic execution path failed to align the time units with the specific problem constraints (likely over-scaling the time duration), leading to a large deviation. \\
\midrule
S+Tools (TTE) &
Predicted $31.6~\mathrm{g}$ and $1.28 \times 10^{3}~\mathrm{cm^2}$ (correct).
Decomposition enforced a step-by-step validation. The system evolved specific tools for electron mole calculation and area derivation, ensuring dimensional consistency at each interface. \\
\bottomrule
\end{tabularx}
\caption{Execution results for Case 2. Step-by-step decomposition prevents error propagation in multi-stage physics problems.}
\label{tab:case_electroplating}
\end{table*}

\begin{table*}[h]
\centering
\resizebox{\textwidth}{!}{%
\begin{tabular}{c l l l l}
\toprule
\textbf{Step} & \textbf{Sub-Goal Description} & \textbf{Tool Action} & \textbf{Tool Status} & \textbf{Intermediate Result} \\
\midrule
1 & Calculate total charge $Q = I \times t$ & \texttt{calculate\_charge(8.46, ...)} & \textcolor{blue}{Retrieved} & $30,500~C$ \\
\rowcolor{gray!12} 2 & Calculate moles of electrons $n = Q/F$ & \texttt{calculate\_moles\_of\_electrons(30500)} & \textcolor{red}{\textbf{Evolved}} & $0.316~mol$ \\
3 & Stoichiometry (Ag oxidation +1) & [Logic Reasoning] & - & $0.316~mol~Ag$ \\
4 & Convert moles to mass (Molar Mass) & \texttt{calculate\_moles(mass, ...)} & \textcolor{blue}{Retrieved} & $31.63~g$ \\
\rowcolor{gray!12} 5 & Calculate volume $Volume = mass / \text{density}$ & \texttt{calculate\_volume(31.63,10.5)} & \textcolor{red}{\textbf{Evolved}} & $3.26~cm^3$ \\
\rowcolor{gray!12} 6 & Calculate tray area $A = V / \text{thickness}$ & \texttt{calculate\_area(3.26, 0.00254)} & \textcolor{red}{\textbf{Evolved}} & $\mathbf{1283.5}~cm^2$ \\
\bottomrule
\end{tabular}%
}
\caption{Step-by-step execution trace for Case 2. The system evolves specific physics and geometry tools (Steps 2 and 6) when exact matches are missing, while reusing standard chemical tools (Step 4) where appropriate.}
\label{tab:case2_trace}
\end{table*}

\begin{figure*}[t]
\centering
\begin{mycodebox}
def calculate_area(volume_cm3, thickness_cm):
    """
    Calculate the area of the tray.

    Args:
        volume_cm3 (float): Volume of silver (cm^3)
        thickness_cm (float): Thickness of the silver coating (cm)

    Returns:
        float: Area (cm^2)
    """
    # Area = Volume / Thickness
    area = volume_cm3 / thickness_cm
    return area
\end{mycodebox}
\caption{Excerpt of a synthesized atomic function for Step 6 of Case 2. The tool was evolved on-the-fly to link the electrochemical result (volume) with the geometric requirement (area).}
\label{fig:case_area_code}
\end{figure*}

\section{Dataset Comparison and Uniqueness}
\label{app:dataset_comparison}
SciEvo fills a critical gap in current evaluation protocols by establishing a benchmark that simultaneously assesses scientific reasoning accuracy and the validity of the tool evolution process. As summarized in Table~\ref{tab:dataset_comparison}, existing benchmarks typically isolate these capabilities, whereas SciEvo couples them to simulate the open-ended nature of real-world scientific research.

\subsection{Comparison with Existing Benchmarks}
Current benchmarks can be categorized into three groups, none of which fully capture the test-time evolution paradigm.
First, scientific reasoning benchmarks like SciBench \citep{wang2023scibench} and SciEval \citep{sun2024scieval} focus on problem-solving but assume a fixed setting. SciBench provides problem sets without executable tools, forcing models to rely on internal parametric knowledge or external calculators without a unified interface. SciEval offers multi-level evaluation but lacks a mechanism to assess tool generation.
Second, function calling benchmarks such as ToolBench \citep{qin2023toolllm}, API-Bank \citep{li2023api}, and BFCL \citep{patil2024gorilla} focus on the retrieval and invocation of static libraries. While recent works like CONFETTI \citep{alkhouli2025confetti} and NESTFUL \citep{basu2025nestful} explore complex nested calls, and LongFuncEval \citep{kate2025longfunceval} assesses long-context retrieval, they all operate under the ``Closed-World'' assumption where the toolset is immutable.
Third, code generation benchmarks, \textit{e.g.}, HumanEval \citep{peng2024humaneval} focus on generating standalone code snippets. While they involve synthesis, they do not evaluate the generated code as reusable library components (tools) that must be maintained and retrieved for future tasks.
SciEvo uniquely integrates these dimensions, requiring the agent to not only solve scientific problems but also to maintain and evolve a persistent library of atomic primitives.

\subsection{Domain Coverage and Tool Modality}
A distinct advantage of SciEvo is its comprehensive disciplinary coverage. Previous domain-specific agents rely on manually curated, static toolkits.
For instance, ChemCrow \citep{bran2023chemcrow} provides a specialized library of approximately 19 tools, categorized into general inference (4), molecule manipulation (8), safety checks (3), and reaction processing (4). Similarly, CheMatAgent \citep{wu2025chemagent} expands this to the materials domain, offering about 34 chemistry tools (\textit{e.g.}, molar mass, solution concentration) and 95 materials science tools (\textit{e.g.}, crystal structure analysis, phase diagram calculation).
However, these libraries are static and domain-confined. They lack support for fundamental Mathematics and Physics, which are ubiquitous in interdisciplinary research. As shown in our analysis, SciEvo encompasses not only Chemistry and Materials but also fills the void in Math and Physics. Crucially, unlike the static definitions in ChemCrow or CheMatAgent, the tools in SciEvo are dynamically synthesized, which means the library coverage is theoretically unbounded, capable of evolving from basic arithmetic to complex thermodynamic simulations depending on the inference trajectory.

\begin{table*}[t]
\centering
\resizebox{\textwidth}{!}{%
\begin{tabular}{l c c c c c}
\toprule
Dataset & Domain & Tool Modality & Evolution & Reasoning Type & Tool Coverage \\
\midrule
SciBench \citep{wang2023scibench} & Science (Gen) & None & No & Chain-of-Thought & - \\
SciEval \citep{sun2024scieval} & Science (Gen) & None & No & Multi-level & - \\
ToolBench \citep{qin2023toolllm} & General & Static API & No & API Calling & General APIs \\
BFCL \citep{patil2024gorilla} & General & Static Function & No & AST Parsing & Python/Java/JS \\
ChemCrow \citep{bran2023chemcrow} & Chemistry & Static Library & No & ReAct & ~19 (Chem) \\
CheMatAgent \citep{wu2025chemagent} & Chem \& Mat & Static Library & No & ReAct & ~130 (Chem+Mat) \\
HumanEval \citep{peng2024humaneval} & CS/Coding & One-off Code & No & Code Gen & - \\
\rowcolor{gray!12} \textbf{SciEvo (Ours)} & \textbf{Science (Gen)} & \textbf{Dynamic/Evolved} & \textbf{Yes} & \textbf{TTE-Reasoning} & \textbf{925} \\
\bottomrule
\end{tabular}%
}
\caption{Comparison of SciEvo with existing benchmarks. SciEvo is the only framework that supports Test-Time Evolution across multiple scientific domains, whereas others rely on static libraries or focus solely on code generation without library management.}
\label{tab:dataset_comparison}
\end{table*}

\section{Theoretical Analysis}
\label{sec:theoretical_analysis}

This section provides a rigorous formalization of the mechanisms underpinning the Test-Time Tool Evolution framework. We derive theoretical bounds for tool reusability, analyze the impact of library scaling on retrieval fidelity, quantify error propagation in sequential reasoning, and prove the convergence of the library size under our pruning strategy.

\subsection{Utility Gain from Atomic Decomposition}
\label{sec:theory_atomic}

We analyze the expected utility of decomposing a monolithic tool into atomic functions. Let $\mathcal{Q}$ be the space of all possible future queries. A monolithic tool $T$ is defined as a composition of $k$ independent atomic operations $\{a_1, a_2, \dots, a_k\}$. For any query $q \in \mathcal{Q}$, let $S(q)$ denote the set of atomic operations required to solve $q$.

\paragraph{Definitions.}
We define the applicability of a tool using indicator variables. The monolithic tool $T$ is applicable to query $q$ if and only if the query requires the complete set of operations implemented by $T$, or a set sufficiently similar that $T$ is retrieved and executed monolithically. For strict analysis, we assume $T$ is reused if $S(q) \supseteq \{a_1, \dots, a_k\}$. Conversely, an atomic tool $A_i$ (corresponding to operation $a_i$) is reused if $a_i \in S(q)$. Let $R(T)$ and $R(A_i)$ be random variables representing the reuse counts of the monolithic tool and the $i$-th atomic tool over a stream of $M$ queries.

\paragraph{Theorem 1 (Decomposition Lower Bound).}
Let $p_{\text{partial}}$ be the probability that a query requires a proper subset of operations, specifically that for any $i$, $P(a_i \in S(q) \mid \{a_1 \dots a_k\} \not\subseteq S(q)) \ge \delta$ for some $\delta > 0$. The expected aggregate reuse of the decomposed atomic library strictly exceeds that of the monolithic tool:
\begin{equation}
\mathbb{E}\left[\sum_{i=1}^{k} R(A_i)\right] \ge k \cdot \mathbb{E}[R(T)] + \Delta_{\text{flexibility}},
\end{equation}
where $\Delta_{\text{flexibility}}$ is a positive term representing the utility of partial reuse.

\paragraph{Proof.}
Consider a single query $q$. Let $X_T^{(q)}$ be the indicator that $T$ is reused, and $X_i^{(q)}$ be the indicator that $A_i$ is reused. By definition, if $T$ is reused, all underlying operations are active, implying $X_T^{(q)} = 1 \implies \forall i, X_i^{(q)} = 1$. Therefore, $X_i^{(q)} \ge X_T^{(q)}$ almost surely.
Taking expectations over the query distribution, we have $\mathbb{E}[X_i^{(q)}] = P(a_i \in S(q))$ and $\mathbb{E}[X_T^{(q)}] = P(\{a_1 \dots a_k\} \subseteq S(q))$.
By the law of total probability, we expand the atomic reuse probability:
\begin{equation}
\begin{split}
P(a_i \in S(q)) &= P(a_i \in S(q) \mid \text{Mono})P(\text{Mono}) \\
\quad &+ P(a_i \in S(q) \mid \text{Partial})P(\text{Partial}),
\end{split}
\end{equation}
Since $P(a_i \in S(q) \mid \text{Mono}) = 1$, the first term is exactly $\mathbb{E}[X_T^{(q)}]$. The second term represents the ``partial match'' scenario where the monolithic tool fails but the atomic tool succeeds. Summing over all $k$ tools and all $M$ queries, and applying the linearity of expectation, we obtain:
\begin{equation}
\begin{split}
\sum_{i=1}^{k} \mathbb{E}[R(A_i)] &= M \sum_{i=1}^{k} \Big( \mathbb{E}[X_T^{(q)}] \\
&\quad + P(a_i \in S(q), \text{Partial}) \Big).
\end{split}
\end{equation}
The term $M \sum \mathbb{E}[X_T^{(q)}]$ equals $k \cdot \mathbb{E}[R(T)]$. The remaining sum is strictly positive given that $p_{\text{partial}} > 0$ and the operations have non-zero marginal utility. Thus, decomposition guarantees higher expected reusability by capturing the marginal utility of sub-problems that monolithic tools miss. \qed

\subsection{Retrieval Precision in Growing Libraries}
\label{sec:theory_retrieval}

We formally examine the ``Tool Overload'' phenomenon. Consider a library of size $N$. For a given query, let $s_r$ be the similarity score of the unique relevant tool, drawn from a distribution with PDF $f_r(s)$ and CDF $F_r(s)$. Let $\{s_{i}\}_{i=1}^{N-1}$ be the scores of $N-1$ irrelevant tools (distractors), independently drawn from a noise distribution with PDF $f_n(s)$ and CDF $F_n(s)$. The retrieval system selects the tool with the maximum score.

\paragraph{Theorem 2 (Monotonic Degradation).}
Assuming the support of the relevant and noise distributions overlap such that $F_n(s) < 1$ for some $s$ where $f_r(s) > 0$, the probability of correctly retrieving the relevant tool, denoted as $P_N(\text{success})$, is a strictly decreasing function of the library size $N$.

\paragraph{Proof.}
The relevant tool is retrieved if its score $s_r$ is greater than the maximum of all $N-1$ distractor scores. Let $M_{N-1} = \max\{s_{1}, \dots, s_{N-1}\}$. The CDF of the maximum of independent variables is the product of their CDFs, so $P(M_{N-1} \le x) = [F_n(x)]^{N-1}$.
The probability of success is the probability that $s_r > M_{N-1}$. We integrate over all possible values of $s_r$:
\begin{equation}
\begin{split}
P_N(\text{success}) &= \int P(M_{N-1} < s \mid s_r = s) f_r(s) \, ds \\
&= \int [F_n(s)]^{N-1} f_r(s) \, ds.
\end{split}
\end{equation}
To determine the trend with respect to $N$, we treat $N$ as a continuous variable and differentiate under the integral sign using Leibniz's rule:
\begin{equation}
\frac{\partial}{\partial N} P_N(\text{success}) = \int_{-\infty}^{\infty} f_r(s) \frac{\partial}{\partial N} [F_n(s)]^{N-1} \, ds.
\end{equation}
Calculating the derivative inside the integral:
\begin{equation}
\frac{\partial}{\partial N} [F_n(s)]^{N-1} = [F_n(s)]^{N-1} \ln(F_n(s)).
\end{equation}
Since $F_n(s)$ is a cumulative distribution function, $0 \le F_n(s) \le 1$, which implies $\ln(F_n(s)) \le 0$. For any region where the distributions overlap and retrieval is non-trivial, $F_n(s) < 1$, making the logarithm strictly negative. Thus, the integrand is non-positive everywhere and strictly negative on the set of overlap. Consequently, $\frac{\partial P_N}{\partial N} < 0$, proving that expanding the library without enhancing the retrieval mechanism (\textit{e.g.}, via sub-question decomposition) inevitably increases the error rate.

\subsection{Stability of Library Growth}
\label{sec:theory_convergence}

We model the temporal dynamics of the tool library size $L(t)$ to prove that the proposed evolution mechanism leads to a stable system rather than unbounded growth.

\paragraph{Dynamics Model.}
The rate of change in library size is governed by two opposing forces: the generation of new tools upon retrieval failure and the pruning of low-utility tools.
Let $\lambda_g$ be the maximum potential generation rate. As the library grows, the probability of finding a match increases, suppressing new generation. We model this saturation with a logistic term $(1 - L(t)/K)$, where $K$ represents the effective capacity of the semantic space.
Let $\lambda_p$ be the pruning rate, proportional to the current library size (assuming a constant fraction of tools falls below the usage threshold).
The differential equation describing the system is:
\begin{equation}
\frac{dL}{dt} = \lambda_g \left(1 - \frac{L(t)}{K}\right) - \lambda_p L(t).
\end{equation}

\paragraph{Theorem 3 (Convergence).}
For any non-negative initial condition $L(0) \ge 0$, the library size $L(t)$ converges asymptotically to a stable equilibrium point $L^*$.

\paragraph{Proof.}
We rearrange the differential equation into a standard linear form with constant coefficients:
\begin{equation}
\frac{dL}{dt} = \lambda_g - \left( \frac{\lambda_g}{K} + \lambda_p \right) L(t).
\end{equation}
Let $A = \lambda_g$ and $B = \frac{\lambda_g}{K} + \lambda_p$. The equation becomes $\frac{dL}{dt} = A - B L(t)$.
The equilibrium point is found by setting $\frac{dL}{dt} = 0$, yielding $L^* = \frac{A}{B} = \frac{\lambda_g K}{\lambda_g + \lambda_p K}$.
The general solution to this first-order linear ordinary differential equation is:
\begin{equation}
L(t) = L^* + \left( L(0) - L^* \right) e^{-B t}.
\end{equation}
Since $B > 0$ (as generation rate, capacity, and pruning rate are all positive), the term $e^{-B t}$ decays to zero as $t \to \infty$. Therefore, regardless of whether the library starts empty or full, the system autonomously regulates itself towards the steady-state size $L^*$. This proves that the TTE framework is robust against explosion in library size, ensuring long-term computational efficiency.

\section{Future Directions and Broader Impact}
\label{sec:future_work}

The transition to evolutionary tool ecosystems offers a generalizable paradigm for intelligence in non-stationary environments. By treating tools as adaptive capabilities rather than static resources, the Test-Time Tool Evolution framework enables agents to navigate open-ended challenges. We outline key directions to scale and robustify this paradigm.

\paragraph{Lifecycle Management.}
Unbounded library growth demands rigorous maintenance to preserve retrieval efficiency. Future research must address the trade-off between plasticity (acquiring new tools) and stability (retaining core competencies). Mechanisms such as intelligent pruning and hierarchical indexing will be critical to forget obsolete primitives while consolidating high-utility functions, preventing knowledge saturation.

\paragraph{Robustness and Calibration.}
Enhancing the reliability of synthesized tools is a priority. Future systems should incorporate formal verification or uncertainty-aware generation to guarantee code safety. Furthermore, we envision meta-cognitive calibration, where agents dynamically weight the cost of retrieval versus evolution based on confidence, alongside self-correction loops that refine tool logic iteratively upon execution failure.

\paragraph{Multi-Modal Frontiers.}
Real-world problems require interpreting diagrams or instrument readouts. Extending TTE to multi-modal contexts involves evolving tools for vision-based analysis or graph manipulation. Co-evolving perception and reasoning capabilities represents a key step toward fully autonomous agents capable of conducting end-to-end scientific research.
\end{document}